\documentclass[review]{elsarticle}

\usepackage{lineno,hyperref}

\usepackage{algorithm}

\usepackage{microtype}
\usepackage{graphicx}
\usepackage{booktabs} 
\usepackage{times}
\usepackage{epsfig}
\usepackage{graphicx}
\usepackage{amsmath}
\usepackage{amssymb}
\usepackage{cite}
\usepackage{wrapfig}
\usepackage[utf8]{inputenc} 
\usepackage[T1]{fontenc}    
\usepackage{hyperref}       
\usepackage{url}            
\usepackage{booktabs}       
\usepackage{amsfonts}       
\usepackage{nicefrac}       
\usepackage{microtype}      
\usepackage{xcolor}         
\usepackage[noend]{algpseudocode}
\usepackage{color}
\usepackage{amsthm}
\usepackage{mwe}
\usepackage{float}
\usepackage{subfig}
\usepackage{tikz}

\usepackage{thmtools, thm-restate}
\theoremstyle{plain}

\usepackage[font=small,labelfont=bf]{caption}

\usepackage{booktabs}
\usepackage{multirow}
\usepackage{cite}
\usepackage[normalem]{ulem}

\definecolor{pink}{rgb}{0.858, 0.188, 0.478}

\newcommand{\cmmnt}[1]{\ignorespaces}

\modulolinenumbers[5]

\journal{Pattern Recognition}

\begin{document}

\begin{frontmatter}
\title{ScanMix: Learning from Severe Label Noise via Semantic Clustering and Semi-Supervised Learning}

\author[vgg,pa_ragav]{Ragav Sachdeva\corref{mycorrespondingauthor}}
\cortext[mycorrespondingauthor]{Corresponding author}
\ead{removethisifyouarehuman-rs@robots.ox.ac.uk}
\fntext[pa_ragav]{\textit{Present Address:} Visual Geometry Group, Department of Engineering Science, University of Oxford, United Kingdom}

\author[ufrpe]{Filipe Rolim Cordeiro}
\author[Magdeburg]{Vasileios Belagiannis}
\author[adelaide]{Ian Reid}
\author[adelaide,surrey]{Gustavo Carneiro}

\address[vgg]{Visual Geometry Group, Department of Engineering Science, University of Oxford, United Kingdom}
\address[adelaide]{School of Computer Science, Australian Institute for Machine Learning, Australia}
\address[ufrpe]{Visual Computing Lab, Department of Computing, Universidade Federal Rural de Pernambuco, Brazil}
\address[Magdeburg]{Otto-von-Guericke-Universit\"at Magdeburg, Germany}
\address[surrey]{Centre for Vision, Speech and Signal Processing, University of Surrey, United Kingdom}

\begin{abstract}
We propose a new training algorithm, ScanMix, that explores semantic clustering and semi-supervised learning (SSL) to allow superior robustness to severe label noise and competitive robustness to non-severe label noise problems, in comparison to the state of the art (SOTA) methods.
ScanMix is based on the expectation maximisation framework, where the E-step estimates the latent variable to cluster the training images based on their appearance and classification results, and the M-step optimises the SSL classification and learns effective feature representations via semantic clustering.
We present a theoretical result that shows the correctness and convergence of ScanMix, and an empirical result that shows that ScanMix has SOTA results on CIFAR-10/-100 (with symmetric, asymmetric and semantic label noise), Red Mini-ImageNet (from the Controlled Noisy Web Labels), Clothing1M and WebVision. 
In all benchmarks with severe label noise, our results are competitive to the current SOTA.
\end{abstract}

\begin{graphicalabstract}
\includegraphics[width=\linewidth]{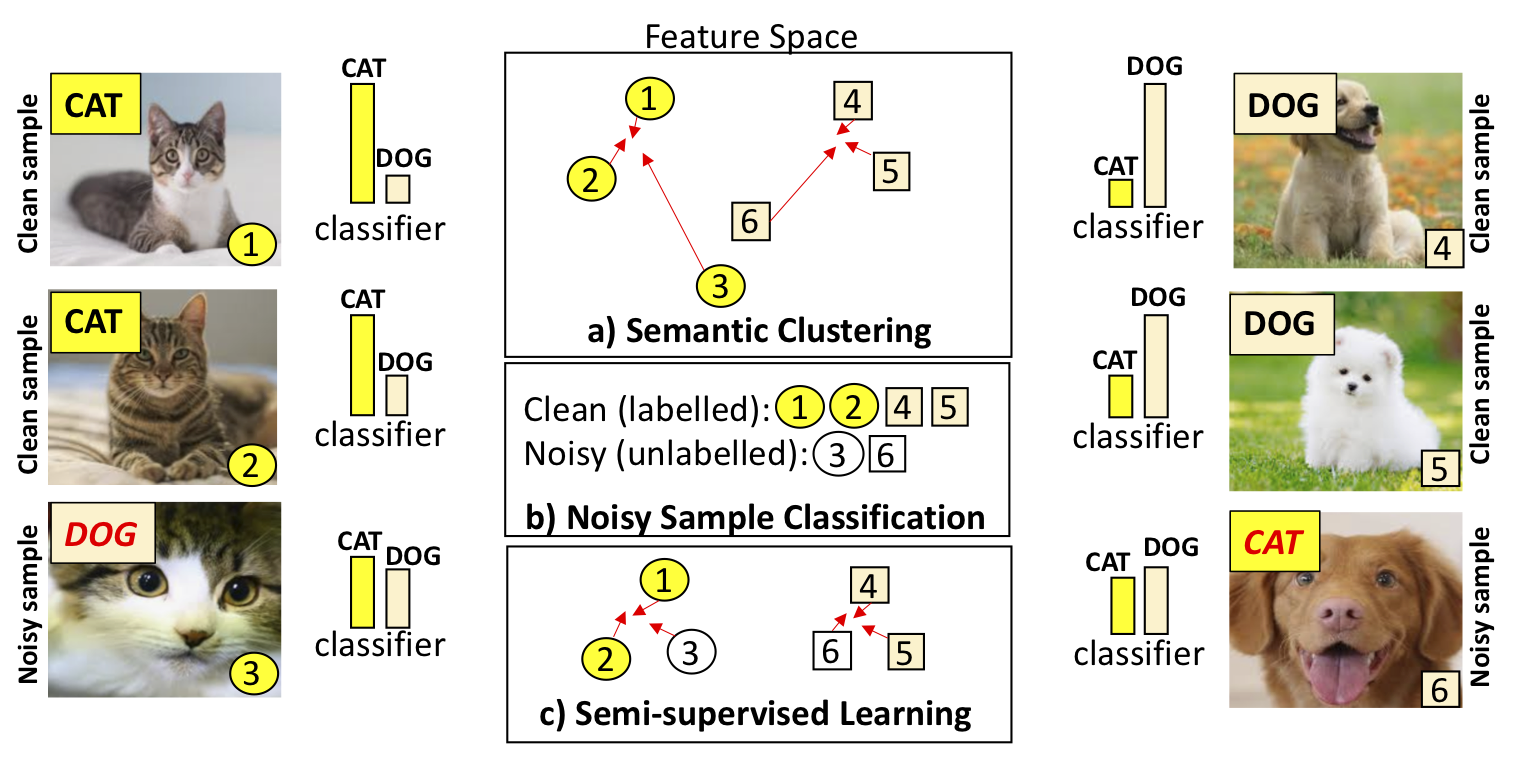}
\end{graphicalabstract}

\begin{highlights}
\item A new noisy-label learning algorithm, called ScanMix
\item ScanMix combines semantic clustering and semi-supervised learning
\item ScanMix is remarkably robust to severe label noise rates
\item  ScanMix provides competitive performance in a wide range of noisy-label learning problems
\item  A new theoretical result that shows the correctness and convergence of ScanMix
\end{highlights}

\begin{keyword}
Noisy label learning \sep Semi-supervised learning \sep Semantic clustering \sep Self-supervised Learning \sep  Expectation maximisation
\end{keyword}

\end{frontmatter}

\linenumbers

\section{Introduction}
\label{sec:introduction}

Much of the success of deep learning models is attributable to the availability of 
well-curated large-scale datasets that enables a reliable supervised learning process~\citep{litjens2017survey}.
However, the vast majority of real-world datasets have noisy labels due to human failure, poor quality of data or inadequate labelling process~\citep{frenay_survey}.
Using noisy label datasets for training not only hurts the model's accuracy, but also biases the model to make the same mistakes present in the labels~\citep{Zhang2017UnderstandingDL}.
Therefore, one of the important challenges in the field is the formulation of robust training algorithms that work effectively with datasets corrupted with noisy labels.

Successful approaches to address the learning from noisy label (LNL) problem tend to rely on semi-supervised learning (SSL)~\citep{ding2018semi,ortego2019towards,ortego2020multi,li2020dividemix}. Such methods run the following steps iteratively: a) automatically split the training set into clean and noisy sets, b) discard the labels of the samples in the noisy set and, c) minimise the classification loss with the labelled (clean) and unlabelled (noisy) data. 
Consequently, these SSL methods rely on successfully splitting the training set into clean and noisy sets, which for low noise rates, is  accurate~\citep{ding2018semi,ortego2019towards,ortego2020multi,li2020dividemix} because of the strong support in the training set that associates image representations and their true labels. However, for severe label noise, this support weakens, resulting in the over-fitting of label noise~\citep{ding2018semi,ortego2019towards,ortego2020multi,li2020dividemix}.

\begin{figure}[t]
\centering
  \includegraphics[width=0.9\linewidth]{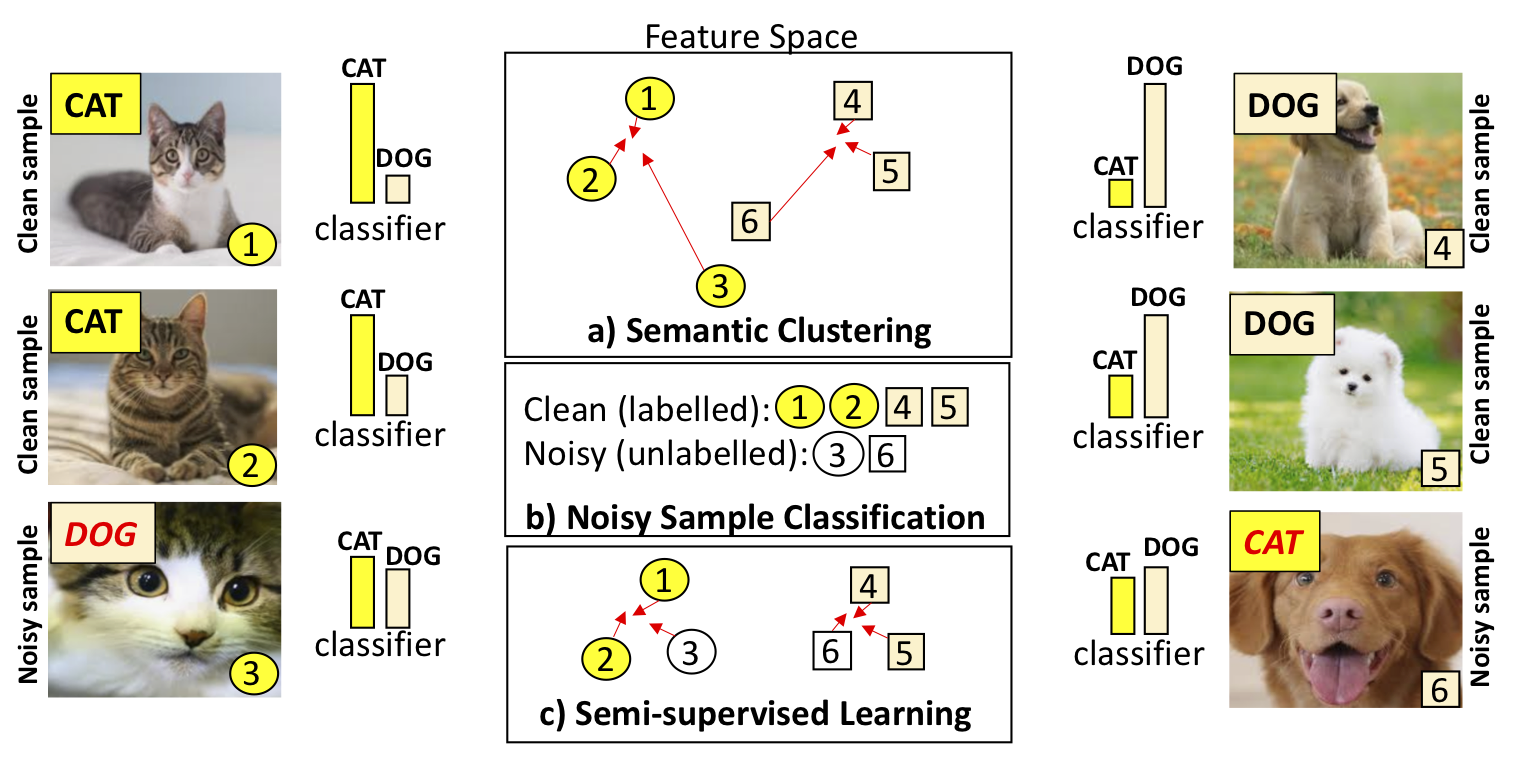}
  \caption{\textbf{ScanMix} explores \textbf{semantic clustering} (a) that clusters samples with similar appearances and classification results, and \textbf{semi-supervised learning (SSL)} (c) that trains the classifier by treating the samples classified to have noisy labels, as unlabelled samples.  In the figure, circles represent true cat label, and squares, true dog class, where samples 3 and 6 are noisy, but the classifier produces the right classification (see yellow and pink bars). In frames (a),(c) the arrows denote how the training process moves samples in the feature space at each stage, with samples 3 and 6 showing white background in (b),(c) because they are classified as noisy in (b) and have their labels removed for SSL.}
  \label{fig:motivation}
\end{figure}

To mitigate the issues caused by severe label noise, one can consider self-supervised learning strategies~\citep{MoCo,MoCoV2,SCAN,SimCLR} to build feature representations 
using appearance clustering techniques. 
These strategies~\citep{MoCo,MoCoV2,SCAN,SimCLR} show better classification accuracy than recently proposed LNL methods~\citep{li2020dividemix,SELF} when the noise rate is large (above 80\% symmetric and 40\% asymmetric). However, for low noise rates, SSL methods tend to produce better results because self-supervised methods typically tend to cluster images with similar appearance, but such similarity does not imply that the images belong to same class. 
We argue that in a noisy label context, the use of self-supervised learning (without using the training set labels) \textit{can} create an initial feature representation that is more related to the real hidden representation \textit{in comparison to} supervised training with noisy labels. However, when the dataset is relatively well-structured and clean, the use of self-supervised learning alone is not enough to bridge the gap to its supervised training counterpart. 
To this end, we propose a training mechanism that performs semantic clustering and SSL in tandem. 
We hypothesise that such training  1) enables the model to not get too biased by noisy labels (as it is guided by semantic clustering), and 2) still produces accurate classification results using the SSL strategy. These points are illustrated in Fig.~\ref{fig:motivation}.

We test this hypothesis with the proposed label-noise robust training algorithm, ScanMix, that explores semantic clustering and SSL to 
enable superior robustness to severe label noise and competitive robustness to non-severe label noise problems, compared with the state of the art (SOTA).
ScanMix is based on the expectation maximisation (EM) framework~\citep{dempster1977maximum}, where the semantic clustering stage clusters images with similar appearance \textit{and} classification results, enabling a more effective identification of noisy label images to subsequently be ``unlabelled" and used by SSL.
Although the use of EM in the context of LNL has been explored in~\citep{rebbapragada2007class}, ScanMix is the first to propose the joint exploration of semantic clustering and SSL together.
The implementation of ScanMix relies on SOTA semantic clustering~\citep{SCAN} and noisy label robust SSL~\citep{li2020dividemix}.
We show a theoretical result that proves that ScanMix is correct and converges to a stationary point under certain conditions. The main contributions of ScanMix are:
\begin{itemize}
    \item A new noisy-label learning algorithm, based on EM optimisation, that explores and combines the advantages of semantic clustering and semi-supervised learning showing remarkable robustness to severe label noise rates; 
    \item A new theoretical result that shows the correctness and convergence of the noisy-label learning algorithm; and
    \item Competitive performance in a wide range of noisy-label learning problems, such as symmetric, asymmetric, semantic or instance dependent, and controlled noisy web labels.
\end{itemize}
Empirical results on CIFAR-10/-100~\citep{krizhevsky2009learning} under symmetric, asymmetric and semantic 
noise, show that ScanMix outperforms  previous approaches. 
For high-noise rate problems in CIFAR-10/-100~\citep{li2020dividemix} and 
Red Mini-ImageNet from the Controlled Noisy Web Labels~\citep{FaMUS}, ScanMix presents the best results in the field. 
Furthermore, we show results on the challenging semantic label noise present in the large-scale real-world datasets Clothing1M~\citep{xiao2015learning} and WebVision~\citep{li2017webvision}, where our proposed method shows SOTA results.

\section{Prior Work}
\label{sec:prior_work}

The main noisy label learning techniques are: 
label cleansing \citep{jaehwan2019photometric, yuan2018iterative}, 
iterative label correction~\citep{zhang2021learning}, 
robust loss functions~\citep{liu2020early, wang2019imae, wang2019symmetric}, meta-learning \citep{han2018pumpout,FaMUS,sun2021learning}, sample weighting \citep{ren2018learning}, ensemble learning \citep{miao2015rboost}, student-teacher model~\citep{tarvainen2017mean}, co-teaching~\citep{li2020dividemix,jiang2018mentornet,malach2017decoupling,han2018co,yu2019does}, dimensionality reduction of the image representation~\citep{ma2018dimensionality}, 
and combinations of the techniques above~\citep{yu2018learning, kim2019nlnl, zhang2019metacleaner,SELF,jiang2020beyond}. 
Recent advances showed that the most promising strategies are based on the combination of co-training, noise filtering, data augmentation and SSL~\citep{li2020dividemix,ortego2019towards,ortego2020multi}. 
Below, we do not review approaches that require a clean validation set, such as~\citep{zhang2020distilling}, since that setup imposes a strong constraint on the type of noisy-label learning problem.

Instead, we focus on methods based on SSL for noisy-label training~\citep{ding2018semi,ortego2019towards,li2020dividemix,ortego2020multi,sachdeva2021evidentialmix,cordeiro2021longremix}, which usually show SOTA results on several benchmarks.  These methods rely on: 1) the identification of training samples containing noisy labels and the subsequent removal of their labels; and 2) performing SSL~\citep{MixMatch} using this set of unlabelled samples and the remaining set of labelled samples.
\citep{ding2018semi} identify a small portion of clean samples from the noisy training set by associating them with high confidence. Then, they use the filtered samples as labelled and the remaining ones as unlabelled in an SSL approach. However, relying on highly confident samples to compose the labelled set may not work well for severe noise rate scenarios because this labelled set can be contaminated with high noise rate.
The methods in~\citep{li2020dividemix,ortego2019towards} classify the noisy and clean samples by fitting a two-component Gaussian Mixture Model (GMM) on the normalised loss values for each training epoch. Next, they use  MixMatch~\citep{MixMatch} to combine the labelled and unlabelled sets with MixUp~\citep{zhang2017mixup}. 

However, these strategies do not perform well for high noise rates because MixUp tends to be ineffective in such scenario.
SSL methods can be robust to severe label noise by exploring a feature clustering scheme that pulls together samples that are semantically similar, without considering the noisy labels from the training set, and one way to enable such semantic clustering is provided by self-supervised learning~\citep{SimCLR,MoCo,MoCoV2,SCAN}.

Self-supervised learning has been used as a pre-training approach to estimate reliable features from  unlabelled datasets, but we are not aware of methods that use it for semantic clustering.  For instance, SimCLR~\citep{SimCLR} generates data augmentations of the input images and trains the model to have similar representation of an image and its augmented samples, while increasing the dissimilarity to the other images. 
MoCo~\citep{MoCo, MoCoV2} tackles self-supervised representation learning by conflating contrastive learning with a dictionary look-up. The proposed framework builds a dynamic dictionary with a queue and a moving-averaged encoder to enable building a large and consistent dictionary on-the-fly that facilitates contrastive unsupervised learning.
Another example is SCAN~\citep{SCAN} that
has several stages of self-supervised training: one based on SimCLR~\citep{SimCLR}, followed by another based on a nearest neighbor clustering scheme, and another based on self-labelling.
Self-supervised learning approaches usually show results better than the noisy label SOTA methods for severe label noise problems (above 80\% noise), but for relatively low label noise rates (below 50\% noise), self-supervised learning tends to be worse.
Therefore, the main question we address in this paper is how to explore the semantic clustering capability of self-supervised learning approaches together with SSL methods, to improve the current SOTA results in severe label noise problems, and maintain the SOTA results in low noise label scenarios. 
Even though sophisticated clustering approaches have been proposed in the field~\citep{wang2018detecting,wang2020robust,shukla2020semi,han2020unsupervised}, we opted to use a simple Euclidean-distance based K-nearest neighbour clustering approach.
Furthermore, semantic clustering has been explored in LNL problems~\citep{rebbapragada2007class,chiaroni2019hallucinating}, but without relying on SSL methods.

\section{Method}
\label{sec:method}

\subsection{Dataset and Label Noise Types}
\label{sec:dataset_labelnoisetype}

Let the training set be denoted by $\mathcal{D}=\{(\mathbf{x}_i, \mathbf{y}_i)\}_{i=1}^{|\mathcal{D}|}$, with $\mathbf{x}_i \in \mathcal{S} \subseteq \mathbb R^{H \times W}$ being the $i^{th}$ image, 
and $\mathbf{y}_i \in \{0,1\}^{|\mathcal{Y}|}$ a one-hot vector of the noisy label, where $\mathcal{Y} = \{1,...,|\mathcal{Y}|\}$ represents the  set of labels, and $\sum_{c \in \mathcal{Y}} \mathbf{y}_i(c)=1$. 
The latent true label of the $i^{th}$ training instance is denoted by $\hat{\mathbf{y}}_i \in \mathcal{Y}$, where $\sum_{c \in \mathcal{Y}} \hat{\mathbf{y}}_i(c)=1$.
This latent true label is used by a noise process to produce $\mathbf{y}_i \sim p(\mathbf{y} | \mathbf{x}_i,\mathcal{Y},\hat{\mathbf{y}}_i)$, with $p(\mathbf{y}(j) | \mathbf{x}_i,\mathcal{Y},\hat{\mathbf{y}}_i(c))=\eta_{jc}(\mathbf{x}_i)$,
where $\eta_{jc}(\mathbf{x}_i) \in [0,1]$ and $\sum_{j \in \mathcal{Y}}\eta_{jc}(\mathbf{x}_i)=1$.  

The types of noises considered in this paper are: symmetric~\citep{kim2019nlnl}, asymmetric~\citep{patrini2017making},  
semantic~\citep{rog}, 

and real-world noise~\citep{jiang2020beyond,li2017webvision,xiao2015learning}.
The symmetric (or uniform) noise flips the latent true label $\hat{\mathbf{y}}_i \in \mathcal{Y}$ to any of the labels in $\mathcal{Y}$ (including the true label) with a fixed probability $\eta$, so  $\eta_{jc}(\mathbf{x}_i)=\frac{\eta}{|\mathcal{Y}|-1}, \forall j,c \in \mathcal{Y}, \text{ such that } j \neq c$, and $\eta_{cc}(\mathbf{x}_i)=1-\eta$. 
The asymmetric noise flips the labels between semantically similar classes~\citep{patrini2017making}, so $\eta_{jc}(\mathbf{x}_i)$ is based on a transition matrix between classes $j,c\in\mathcal{Y}$, but not on $\mathbf{x}_i$. 
The semantic noise~\citep{rog} 

also uses an estimated transition probability between classes $j,c\in\mathcal{Y}$ but takes into account the image $\mathbf{x}_i$ (i.e., it is an image conditional transition probability).

Real-world noise~\citep{jiang2020beyond,li2017webvision,xiao2015learning} contains the noise types above in addition to the open-set noise, where the class $c\notin\mathcal{Y}$.

\subsection{ScanMix}
\label{sec:scanmix}

\begin{figure*}
\centering
  \includegraphics[width=1.0\linewidth]{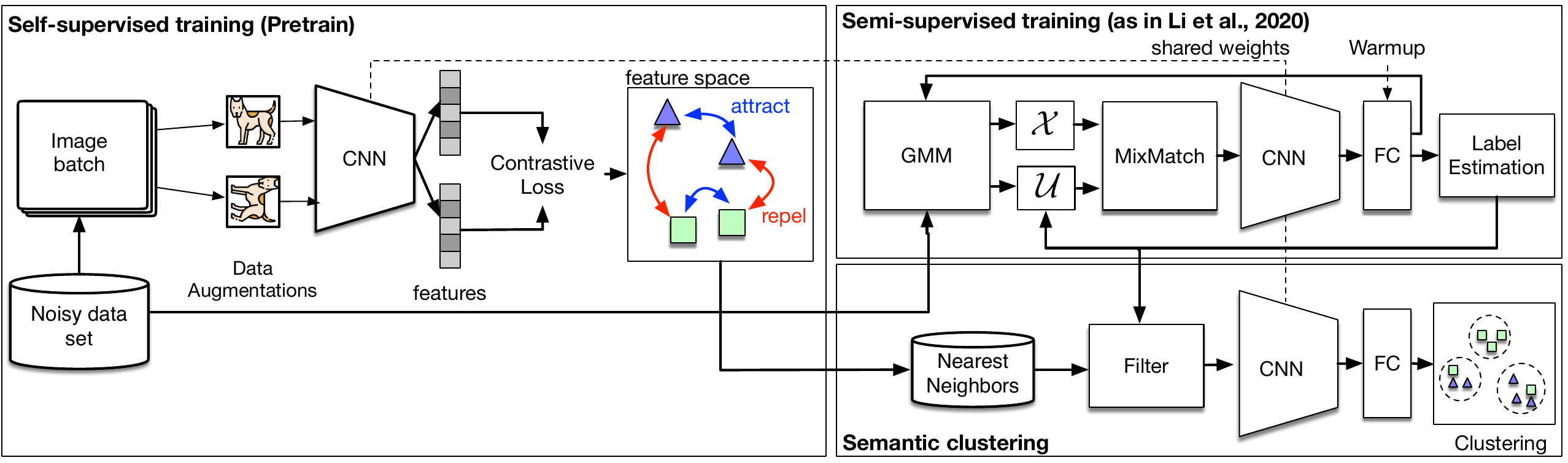}
  \caption{ScanMix has a pre-training stage consisting of a self-supervised training~\citep{SimCLR,MoCo,MoCoV2,SCAN}, where we use contrastive loss to approximate features to its data augmented variants, in the feature space, while repelling representations from negative examples. In the training stage we first warm-up the classifier using a simple classification loss.  Then, using the classification loss, we train the GMM to separate the samples into a clean set $\mathcal{X}$ and a noisy set $\mathcal{U}$ that are "MixMatched"~\citep{MixMatch} for SSL training.  In parallel to this SSL training, we use the classification results and feature representations to train the semantic clustering.  Please see  Algorithm~\ref{alg:SM} for more details.}
  \label{fig:scanmix}
\end{figure*}

The proposed ScanMix training algorithm (Fig.~\ref{fig:scanmix}) is formulated with an EM algorithm that uses a latent random variable $z_{ji} \in \{0,1\}$ which indicates if a sample $\mathbf{x}_j$ belongs to the set of $K$ nearest neighbours (KNN) of $\mathbf{x}_i$, estimated with the Euclidean distance.

The classifier trained by ScanMix is parameterised by $\theta = [ \psi, \phi ] \in \Theta$, and represented by
\begin{equation}
p_{\theta}(\mathbf{y}|\mathbf{x}) = p_{\psi}(\mathbf{y} | f_{\phi}(\mathbf{x})),
\label{eq:classifier}    
\end{equation}
where $p_{\psi}(.) \in [0,1]^{|\mathcal{Y}|}$ produces a probability distribution over the classes in the classification space $\mathcal{Y}$ using the feature representation $f_{\phi}(\mathbf{x}) \in \mathbb{R}^d$ of the input image $\mathbf{x}$.

The optimal parameters for the classifier are estimated with maximum likelihood estimation (MLE):
\begin{equation}
    \theta^* = \arg\max_{\theta}  \frac{1}{|\mathcal{D}|}\sum_{(\mathbf{x}_i,\mathbf{y}_i) \in \mathcal{D}} \log p_{\theta}(\mathbf{y}_i|\mathbf{x}_i),
    \label{eq:main_likelihood}
\end{equation}
where
\begin{equation}
\begin{split}
    \log p_{\theta}(\mathbf{y}_i|\mathbf{x}_i) =& \mathbb{E}_{q(z)} \left [ \log \left ( p_{\theta}(\mathbf{y}_i|\mathbf{x}_i)\frac{q(z)}{q(z)} \right ) \right ] 
    \\
    =&\int q(z) \log\left( \frac{p_{\theta}(\mathbf{y}_i,z|\mathbf{x}_i)q(z)}{p_{\theta}(z|\mathbf{y}_i,\mathbf{x}_i)q(z)} \right)dz \\
    =&\mathbb{E}_{q(z)}[\log(p_{\theta}(\mathbf{y}_i,z|\mathbf{x}_i))] -\mathbb{E}_{q(z)}[\log q(z) ] + KL[q(z)||p_{\theta}(z|\mathbf{y}_i,\mathbf{x}_i)] \\
    =& \ell_{ELBO}(q,\theta)+KL[q(z)||p_{\theta}(z|\mathbf{y}_i,\mathbf{x}_i)].
\end{split}
\label{eq:log_likelihood_p_y_x_theta}
\end{equation}
In~Eq.\ref{eq:log_likelihood_p_y_x_theta} above, we have:
\begin{equation}
\ell_{ELBO}(q,\theta) = \mathbb{E}_{q(z)}[\log p_{\theta}(\mathbf{y},z|\mathbf{x})]-\mathbb{E}_{q(z)}[\log q(z) ],   
\label{eq:elbo}
\end{equation}
 with $KL[\cdot]$ denoting the Kullback-Leibler divergence, and $q(z)$ representing the variational distribution that approximates $p_{\theta}(z|\mathbf{y},\mathbf{x})$, defined as
\begin{equation}
\scalebox{0.9}{
$p_{\theta} (z_{ji}|  \mathbf{x}_i  , \mathbf{y}_i) = \left \{
\begin{array}{ll}
  (1 - z_{ji}) & \text{, if }  \mathbf{y}_j \neq  \mathbf{y}_i\\
  (p_{\theta}(:|\mathbf{x}_j)^{\top}p_{\theta}(:|\mathbf{x}_i))^{z_{ji}} 
      (1-p_{\theta}(:|\mathbf{x}_j)^{\top}p_{\theta}(:|\mathbf{x}_i))^{(1-z_{ji})} & \text{, if }  \mathbf{y}_j =  \mathbf{y}_i\end{array}
\right.$}
    \label{eq:p_z_given_y_x_}
\end{equation}
where 

$p_{\theta}(:|\mathbf{x}) \in [0,1]^{|\mathcal{Y}|}$ is the probability classification for defined in~Eq.\ref{eq:classifier}.

Hence, Eq.\ref{eq:p_z_given_y_x_} defines the
probability of $z_{ij} \in \{0,1\}$, denoting the probability of $\mathbf{x}_j$ to belong to the set of $K$ nearest neighbours (KNN) of $\mathbf{x}_i$. 
In this definition, when their labels are different, or $\mathbf{y}_i \neq \mathbf{y}_j$, the probability of $z_{ij}=1$ is 0 (and consequently, the probability of $z_{ij}=0$ is 1).
Also, when their labels are equal, or $\mathbf{y}_i = \mathbf{y}_j$, the probability of $z_{ij}=1$ depends on the similarity of their classification probabilities denoted by $p_{\theta}(:|\mathbf{x}_j)^{\top}p_{\theta}(:|\mathbf{x}_i)$.

The maximisation of the log likelihood in~Eq.\ref{eq:main_likelihood} follows the EM algorithm~\citep{dempster1977maximum} consisting of two steps. The E-step maximizes the lower bound of~Eq.\ref{eq:log_likelihood_p_y_x_theta} by zeroing the KL divergence with
$q(z_{ji})=p_{\theta^{old}}(z_{ji}|\mathbf{y}_i,\mathbf{x}_i)$, where $\theta^{old}$ denotes the parameter from the previous EM iteration.
Then the M-step maximises $\ell_{ELBO}$ in~Eq.\ref{eq:elbo}, which re-writes~Eq.\ref{eq:main_likelihood} as:
\begin{align*}
    \theta^*  & =  \arg\max_{\theta}  
    \frac{1}{|\mathcal{D}|}  \sum_{(\mathbf{x}_i,\mathbf{y}_i) \in \mathcal{D}} \sum_{j=1}^{|\mathcal{D}|} \sum_{z_{ij}\in \{0,1\}} q(z_{ji}) \log p_{\theta}(z_{ji},\mathbf{y}_i|\mathbf{x}_i)
    \\
    &  =  \arg\max_{\theta}
    \frac{1}{|\mathcal{D}|}  \sum_{(\mathbf{x}_i,\mathbf{y}_i) \in \mathcal{D}} \sum_{j=1}^{|\mathcal{D}|} \sum_{z_{ij}\in \{0,1\}} q(z_{ji}) \log \left(p_{\theta}(z_{ji}|\mathbf{y}_i,\mathbf{x}_i)p_{\theta}(\mathbf{y}_i|\mathbf{x}_i)\right)
\end{align*}
which by noting that
\begin{align*}
\sum_{(\mathbf{x}_i,\mathbf{y}_i) \in \mathcal{D}} \sum_{j=1}^{|\mathcal{D}|} \sum_{z_{ij}\in \{0,1\}} q(z_{ji})\log p_{\theta}(\mathbf{y}_i|\mathbf{x}_i) = \sum_{(\mathbf{x}_i,\mathbf{y}_i) \in \mathcal{D}} \sum_{j=1}^{|\mathcal{D}|} \log p_{\theta}(\mathbf{y}_i|\mathbf{x}_i),
\end{align*}
we have
\begin{equation}
\begin{split}
    \theta^*  =  \arg\max_{\theta}   \frac{1}{|\mathcal{D}|}  \sum_{(\mathbf{x}_i,\mathbf{y}_i) \in \mathcal{D}} \sum_{j=1}^{|\mathcal{D}|}
    \Big ( \log p_{\theta}(\mathbf{y}_i|\mathbf{x}_i) +
     \sum_{z_{ij}\in \{0,1\}} q(z_{ji})\log p_{\theta}(z_{ji}|\mathbf{y}_i,\mathbf{x}_i) \Big ) ,
    \end{split}
    \label{eq:M_step}
\end{equation}
where the term $\mathbb{E}_{q(z)}[\log(q(z))]$ is removed from $\ell_{ELBO}$ since it only depends on the parameter from the previous iteration, $\theta^{old}$.  Hence, ~Eq.\ref{eq:M_step} comprises two terms: 1) the \textbf{classification term} that maximises the likelihood of the label $\mathbf{y}_i$ for sample $\mathbf{x}_i$; and 2) the \textbf{semantic clustering term} that maximises the association between samples that are close in the feature and label spaces, according to $q(z_{ji})$ estimated from the E-step.  

According to the Equations~\ref{eq:p_z_given_y_x_} and~\ref{eq:M_step}, the run-time complexity of ScanMix is quadratic in $|\mathcal{D}|$, making this algorithm impractical for large-scale problems.  Therefore, we approximate both steps by running a self-supervised pre-training process~\citep{SimCLR,MoCo,MoCoV2,SCAN} that forms an initial set of $K$ nearest neighbours in the feature space $f_{\phi}(\mathbf{x})$ for each training sample.  
The set of KNN samples for each sample $\mathbf{x}_i \in \mathcal{D}$ is denoted by $\mathcal{N}_{\mathbf{x}_{i}} = \{ \mathbf{x}_j \}_{j=1}^K$ (for $\mathbf{x}_j \in \mathcal{D}$).
Then, $q(z_{ji})$ is approximated to be equal to 1, when $\mathbf{x}_j \in \mathcal{N}_{\mathbf{x}_{i}}$ and $\mathbf{y}_j=\mathbf{y}_i$, and 0 otherwise. Such an approximation makes the run-time complexity of the E and M steps linear in $\mathcal{D}$\footnote{We tested ScanMix without this approximation and preliminary results show that updating neighbors $\{N_{x_i}\}_{i=1}^{|D|}$ leads to similar results as the ones in this paper, suggesting the validity of our approximation.}. Also, using this approximation to estimate $q(z_{ji})$ in Eq.\ref{eq:p_z_given_y_x_} reduces even more the complexity of the semantic clustering maximisation because we
only consider $q(z_{ji}=1)$ instead of $q(z_{ji}=1)$ and $q(z_{ji}=0)$.

The optimisation of the \textbf{classification term} in~Eq.\ref{eq:M_step} assumes that $\mathcal{D}$ is not noisy, so we modify it to enable learning with a noisy dataset.  This is achieved by maximising a lower bound of that term, as follows~\citep{li2020dividemix}:
\begin{equation}
    \begin{split}
        \text{maximise} & \frac{1}{|\mathcal{X}|}\sum_{(\mathbf{x},\mathbf{y}) \in \mathcal{X}} \log p_{\theta}(\mathbf{y}|\mathbf{x}) \\
        \text{subject to} & \frac{1}{|\mathcal{U}|} \sum_{(\mathbf{x},\mathbf{y}) \in \mathcal{U}}\| \mathbf{y} - p_{\theta}(:|\mathbf{x}) \|_2^2 = 0 \\
        & KL \left [ \pi_{|\mathcal{Y}|} \Bigg | \Bigg | \frac{1}{|\mathcal{X}| + |\mathcal{U}|} \sum_{\mathbf{x} \in (\mathcal{X} \bigcup \mathcal{U})} p_{\theta}(:|\mathbf{x}) \right ]=0,
    \end{split}
    \label{eq:max_likelihood_optimisation}
\end{equation}
where $\mathcal{X}$ and $\mathcal{U}$ represent the
sets of samples extracted from $\mathcal{D}$ automatically classified as clean and noisy, respectively, $p_{\theta}(:|\mathbf{x})$ denotes the classification probability for all classes in $\mathcal{Y}$, $KL[.]$ represents the Kullback Leibler (KL) divergence, and $\pi_{|\mathcal{Y}|}$ denotes a vector of $|\mathcal{Y}|$ dimensions with values equal to $1/|\mathcal{Y}|$.  
The classification of training samples into clean or noisy is first formed with~\citep{arazo2019unsupervised, li2020dividemix,rog, jiang2020beyond}:
\begin{equation}
\begin{split}
    \mathcal{X}'  = & \left \{ (\mathbf{x}_i,\mathbf{y}_i) : p \left ( \text{clean} | \ell_i , \gamma \right ) \ge \tau \right \}, \\
    \mathcal{U}'  = & \left \{ (\mathbf{x}_i,\mathbf{y}_i^{*}) : 
p \left ( \text{clean} | \ell_i , \gamma \right ) < \tau \right \},
\end{split}
\label{eq:clean_noisy_sets}
\end{equation} 
with $\tau$ denoting a threshold to classify a clean sample, $\mathbf{y}_i^{*} = p_{\theta}(:|\mathbf{x}_i)$,
$\ell_i = -\mathbf{y}_i^{\top}\log p_{\theta}(:|\mathbf{x}_i)$, and $p \left ( \text{clean} | \ell_i , \gamma \right )$ being a function that estimates the probability that $(\mathbf{x}_i,\mathbf{y}_i)$ is a clean label sample.  The function $p \left ( \text{clean} | \ell_i , \gamma \right )$ in Eq.\ref{eq:clean_noisy_sets} is a bi-modal Gaussian mixture model (GMM)~\citep{li2020dividemix} ($\gamma$ denotes the GMM parameters), where the component with larger man is the noisy component and the smaller mean is the clean component.
Next, we run SSL~\citep{li2020dividemix}, consisting of a data augmentation to increase the number of samples in $\mathcal{X}'$ and $\mathcal{U}'$, followed by 
MixMatch~\citep{MixMatch} 
that combines samples from both sets to form the sets $\mathcal{X}$ and $\mathcal{U}$, which are used in Eq.\ref{eq:max_likelihood_optimisation}. 
The optimisation in Eq.~\ref{eq:max_likelihood_optimisation} is done with Lagrange multipliers by minimising the loss $\ell_{MLE}= \ell_{\mathcal{X}} + \lambda_u\ell_{\mathcal{U}}  +
\lambda_r\ell_{r}$, where $\ell_{\mathcal{X}}$ represents the (negative) objective function, $\ell_{\mathcal{U}}$ and $\ell_{r}$ denote the two constraints, and $\lambda_u$ and $\lambda_r$ are the Lagrange multipliers.

We constrain the optimisation of the \textbf{semantic clustering term} in~Eq.\ref{eq:M_step} with a 
regulariser~\citep{SCAN} to make it robust to semantic drift~\citep{zhang2019addressing}.
Hence, we maximise a lower bound of 
the semantic clustering term in~Eq.\ref{eq:M_step}, as follows:
\begin{equation}
    \begin{split}
        \text{maximise} & \frac{1}{|\mathcal{D}|}  \sum_{(\mathbf{x}_i,\mathbf{y}_i) \in \mathcal{D}} 
   \sum_{j=1}^{|\mathcal{D}|} \sum_{z_{ij}\in \{0,1\}} q(z_{ji})\log p_{\theta}(z_{ji}|\mathbf{y}_i,\mathbf{x}_i) \\
        \text{subject to} & \sum_{c \in \mathcal{Y}} \mathbb{E}_{\mathbf{x}\sim \mathcal{D}}[p(c|\mathbf{x},\theta)] \log \mathbb{E}_{\mathbf{x}\sim \mathcal{D}}[p(c|\mathbf{x},\theta)]=0,
    \end{split}
    \label{eq:cluster_optimisation}
\end{equation}
where $q(z_{ji}) = 1$ if $\mathbf{x}_i$ and $\mathbf{x}_j$ have the same classification result, i.e.,
$\arg\max_{c \in \mathcal{Y}}p_{\theta}(c | \mathbf{x}_i)=\arg\max_{c \in \mathcal{Y}}p_{\theta}(c | \mathbf{x}_j)$ and $\mathbf{x}_j \in \mathcal{N}_{\mathbf{x}_i}$.
We also use Lagrange multipliers to optimise Eq.~\ref{eq:cluster_optimisation}, where we minimise $\ell_{CLU}= \ell_{\mathcal{N}} + \lambda_e\ell_{e}$, with $\ell_{\mathcal{N}}$ denoting the negative objective function, $\ell_{e}$ representing the constraint, and $\lambda_e$ being the Lagrange multiplier. An interesting point from the optimisations in Eq.~\ref{eq:max_likelihood_optimisation}
and Eq.~\ref{eq:cluster_optimisation} is that the constraints can help mitigate the semantic drift problem~\citep{zhang2019addressing} typically present in under-constrained SSL methods.

\begin{table*}[t!]
\centering
\scalebox{0.7}{
\begin{tabular}{lc|cccc|cc||cccc}
\hline
\multicolumn{2}{c}{dataset} & \multicolumn{6}{c}{CIFAR-10} & \multicolumn{4}{c}{CIFAR-100}\\    
\hline
\multicolumn{2}{c}{Noise type} & \multicolumn{4}{c}{sym.} & \multicolumn{2}{c}{asym.} &  \multicolumn{4}{c}{sym.} \\
\hline
Method/ noise ratio & &  20\% & 50\% & 80\% & 90\% & 40\%& 49\% & 20\% & 50\% & 80\% & 90\% \\
\hline
\multirow{2}{*}{Cross-Entropy \citep{li2020dividemix}}& Best & 86.8 & 79.4 & 62.9 & 42.7 & 85.0 & - & 62.0 & 46.7 & 19.9 & 10.1\\
    & Last & 82.7 & 57.9 & 26.1 & 16.8 & 72.3 & - & 61.8 & 37.3 & 8.8 & 3.5\\
\hline

\multirow{2}{*}{Coteaching+ \citep{yu2019does}} & Best & 89.5 & 85.7 & 67.4& 47.9& - & -&65.6& 51.8 & 27.9 & 13.7\\
    & Last & 88.2 & 84.1 & 45.5& 30.1& - & -&64.1& 45.3 & 15.5 & 8.8\\
\hline
\multirow{2}{*}{MixUp \citep{zhang2017mixup}} &Best & \textbf{95.6} & 87.1 & 71.6& 52.2& - & - & 67.8& 57.3 & 30.8 & 14.6\\
 &Last & 92.3 & 77.3 & 46.7& 43.9& - & - & 66.0& 46.6 & 17.6 & 8.1\\
\hline
\multirow{2}{*}{PENCIL \citep{yi2019probabilistic}}& Best & 92.4 & 89.1 & 77.5& 58.9& 88.5 & - & 69.4& 57.5 & 31.1 & 15.3 \\
  & Last & 92.0 & 88.7 & 76.1& 58.2& 88.1 & - & 68.1& 56.4 & 20.7 & 8.8 \\
\hline
\multirow{2}{*}{Meta-Learning \citep{li2019learning}}&Best & 92.9 & 89.3 & 77.4& 58.7& 89.2 & - & 68.5& 59.2 & 42.4 & 19.5 \\
   &Last & 92.0 & 88.8 & 76.1& 58.3& 88.6 & - & 67.7& 58.0 & 40.1 & 14.3 \\
\hline
\multirow{2}{*}{M-correction \citep{arazo2019unsupervised}}&Best& 94.0 & 92.0 & 86.8& 69.1& 87.4 & - & 73.9& 66.1 & 48.2 & 24.3 \\
   &Last& 93.8 & 91.9 & 86.6& 68.7& 86.3 & - & 73.4& 65.4 & 47.6 & 20.5 \\
\hline
\multirow{2}{*}{MentorMix~\citep{jiang2020beyond}}& Best & \textbf{95.6} & - & 81.0 & - & - & - & \textbf{78.6} & - & 41.2 & - \\
  & Last & - & - & - & - & - & - & - & - & - & - \\
\hline
\multirow{2}{*}{MOIT+~\citep{ortego2020multi}}& Best & 94.1 & - & 75.8 & - & \textbf{93.3} & - & 75.9 & - & 51.4 & - \\
  & Last & - & - & - & - & - & - & - & - & - & - \\
\hline
\multirow{2}{*}{DivideMix \citep{li2020dividemix}}& Best & \textbf{96.1} & \textbf{94.6} & \textbf{93.2} & 76.0& \textbf{93.4} & 83.7* & 77.3 & 74.6 & 60.2 & 31.5 \\
  & Last & \textbf{95.7} & \textbf{94.4} & \textbf{92.9} & 75.4& 92.1 & 76.3* & \textbf{76.9} & 74.2 & 59.6 & 31.0 \\
\hline
\multirow{2}{*}{ELR+~\citep{liu2020early}}& Best & \textbf{95.8} & \textbf{94.8} & \textbf{93.3} & 78.7& \textbf{93.0} & - & \textbf{77.6} & 73.6 & 60.8 & 33.4 \\
  & Last & - & - & - & - & - & - & - & - & - & - \\

\hline
 \multirow{2}{*}{PES~\citep{bai2021understanding}} & Best & \textbf{95.9} & \textbf{95.1} & \textbf{93.1} & - & 77.4 & - & 74.3 & 61.6 & 
 - & - \\
  & Last & - & - & - & - & - & - & - & - & - & - \\
\hline
 \multirow{2}{*}{FSR~\citep{zhang2021FSR}} & Best & \textbf{95.1} & - & 82.8 & - & \textbf{93.6} & - & \textbf{78.7} & - & 
 46.7 & - \\
  & Last & - & - & - & - & - & - & - & - & - & - \\
\hline
 \multirow{2}{*}{DRPL~\citep{ortego2019towards}} & Best & \textbf{94.2} & - & 64.4 & - & \textbf{93.1} & - & 71.3 & - & 
 53.0 & - \\
  & Last & - & - & - & - & - & - & - & - & - & - \\
\hline
\multirow{2}{*}{\textbf{ScanMix (Ours)}}& Best & \textbf{96.0} & \textbf{94.5} & \textbf{93.5} & \textbf{91.0} & \textbf{93.7} & \textbf{88.7} & 77.0 & \textbf{75.7} & \textbf{66.0} & \textbf{58.5}\\
    & Last & \textbf{95.7} & \textbf{93.9} & \textbf{92.6} & \textbf{90.3} & \textbf{93.4} & \textbf{87.1} & \textbf{76.0} & \textbf{75.4} & \textbf{65.0} & \textbf{58.2}\\
\hline
\end{tabular}
}
\caption{Test accuracy (\%) for all competing methods on CIFAR-10 and CIFAR-100 under symmetric and asymmetric noises. Results from related approaches are as presented in~\citep{li2020dividemix}. The results with (*) were produced by locally running the published code provided by the authors. Top methods within $1\%$ are in \textbf{bold}.} 
\label{tab:results_cifar}
\end{table*}

\begin{table*}[t!]
\centering
\scalebox{0.69}{
\begin{tabular}{lcccccc}
\toprule
dataset & \multicolumn{3}{c}{CIFAR-10} & \multicolumn{3}{c}{CIFAR-100}\\    
\midrule

Method/ noise ratio & DenseNet (32\%) & ResNet (38\%) & VGG (34\%) & DenseNet (34\%) & ResNet (37\%) & VGG (37\%) \\
\midrule
D2L + RoG~\citep{rog} &  68.57 & 60.25 & 59.94 & 31.67 & 39.92 & 45.42\\
CE + RoG~\citep{rog} &  68.33 & 64.15 & 70.04 & 61.14 & 53.09 & 53.64\\
Bootstrap + RoG~\citep{rog} &  68.38 & 64.03 & 70.11 & 54.71 & 53.30 & 53.76\\
Forward + RoG~\citep{rog} &  68.20 & 64.24 & 70.09 & 53.91 & 53.36 & 53.63\\
Backward + RoG~\citep{rog} &  68.66 & 63.45 & 70.18 & 54.01 & 53.03 & 53.50\\
DivideMix* \citep{li2020dividemix} &  84.57 & 81.61 & 85.71 & \textbf{68.40} & 66.28 & \textbf{66.84}\\
\textbf{ScanMix (Ours)} &  \textbf{89.70} & \textbf{85.58} & \textbf{89.96} & \textbf{68.44} & \textbf{67.36} & \textbf{67.34} \\
\bottomrule \\
\end{tabular}
}
\caption{Test accuracy (\%) for Semantic Noise. Results from baseline methods are as presented in~\citep{rog}. The results with (*) were produced by locally running the published code provided by the authors. 
Top methods within $1\%$ are in \textbf{bold}.} 
\label{tab:res_semantic}
\end{table*}

\subsection{Training, Inference, Correctness and Convergence Conditions}

\begin{algorithm}
\scriptsize
\caption{ScanMix}\label{alg:SM}
\begin{algorithmic}
\Require $\mathcal{D}$, number of epochs $E$, clean sample threshold $\tau$
\State $f_{\phi}(\mathbf{x})$,$\{ \mathcal{N}_{\mathbf{x}_{i}} \}_{i=1}^{|\mathcal{D}|}$ = PreTrain($\mathcal{D}$) 
\Comment{Self-supervised pre-training}
\State $p_{\theta}(\mathbf{y}|\mathbf{x})$ = WarmUp($\mathcal{D}$,$f_{\phi}(\mathbf{x})$) 
\Comment{Warm Up} 
\While{$e < E$}
    \For{$i = \{1,...,|\mathcal{D}|\}$}
 \State Estimate $p(\text{clean}|\ell_i,\gamma)$, with 
 \State $\ell_i=-\mathbf{y}_i^{\top}\log p_{\theta}(:|\mathbf{x}_i)$
 \EndFor
 \State $\mathcal{X}',\mathcal{U}'$=FormCleanNoisySets($\{p(\text{clean}|\ell_i,\gamma)\}_{i=1}^{|\mathcal{D}|},\tau$) 
 \State $\mathcal{X},\mathcal{U}$=MixMatch($\mathcal{X}',\mathcal{U}'$) 
 \For{$(\mathbf{x}_i,\mathbf{y}_i) \in \mathcal{D}$}
  \Comment{E-step} 
 \State $\tilde{y}_i= \arg\max_{c\in\mathcal{Y}} p_{\theta}(c|\mathbf{x}_i)$
 \State $q(z_{ji}) = 0, \forall j\in\{1,...,|\mathcal{D}|\}$ 
 \For{$\mathbf{x}_j \in \mathcal{N}_{\mathbf{x}_{i}}$}
 \State $\tilde{y}_j=\arg\max_{c\in\mathcal{Y}}p_{\theta}(c|\mathbf{x}_j)$
 \If{$(\tilde{y}_i == \tilde{y}_j)$ }
 \State $q(z_{ji}) =1$
 \EndIf
 \EndFor
 \EndFor
  \State Minimise $\ell_{MLE}$ with 
 $\mathcal{X}$, $\mathcal{U}$, and  
 \State $\ell_{CLU}$ with $\{\mathcal{N}_{\mathbf{x}_i}\}_{i=1}^{|\mathcal{D}|}$, and $\{q(z_{ji}) \}_{i,j=1}^{|\mathcal{D}|}$
 \Comment{M-step}
 \EndWhile
\end{algorithmic}
\end{algorithm}

Algorithm~\ref{alg:SM} describes the training process
that starts with a self-supervised pre-training~\citep{SimCLR,MoCo,MoCoV2,SCAN} which optimises the parameters of the feature extractor $f_{\phi}(\mathbf{x})$ using the unlabelled images of $\mathcal{D}$ and defines the set of KNNs for each training sample $\{ \mathcal{N}_{\mathbf{x}_i} \}_{i=1}^{|\mathcal{D}|}$.
Then, we warm-up the classifier by training it for a few epochs on the (noisy) training dataset using cross-entropy loss. 
Next, we run the EM optimisation using~Eq.\ref{eq:max_likelihood_optimisation}
and~Eq.\ref{eq:cluster_optimisation}.
The inference uses the model in~Eq.\ref{eq:classifier} to classify $\mathbf{x}$.

ScanMix improves $\ell_{ELBO}(q,\theta)$ in Eq.\ref{eq:elbo} instead of improving $p_{\theta}(\mathbf{y}|\mathbf{x})$ in Eq.\ref{eq:main_likelihood}.  Following Theorem 1 in~\citep{dempster1977maximum}, \autoref{thm:correctness} shows the correctness of ScanMix, where an improvement to $\ell_{ELBO}(q,\theta)$ implies an increase to $p_{\theta}(\mathbf{y}|\mathbf{x})$.
Following Theorem 2 in~\citep{dempster1977maximum}, \autoref{thm:convergence} shows the convergence conditions of ScanMix.

\begin{restatable}[]{lemma}{Correctness}
    \label{thm:correctness}
    Assuming that the maximisation of $\ell_{ELBO}$ in Eq.\ref{eq:M_step} estimates $\theta$ that makes 
    $\mathbb{E}_{q(z)}[\log p_{\theta}(\mathbf{y},z|\mathbf{x})] \ge 
    \mathbb{E}_{q(z)}[\log p_{\theta^{old}}(\mathbf{y},z|\mathbf{x})]$, we have that $\left(\log p_{\theta}(\mathbf{y}|\mathbf{x}) - \log p_{\theta^{old}}(\mathbf{y}|\mathbf{x})\right)$ is lower bounded by $\left(\mathbb{E}_{q(z)}[\log p_{\theta}(\mathbf{y},z|\mathbf{x})] - 
    \mathbb{E}_{q(z)}[\log p_{\theta^{old}}(\mathbf{y},z|\mathbf{x})]\right) \ge 0$, with $q(z)=p_{\theta^{old}}(z|\mathbf{y},\mathbf{x})$.
 \end{restatable}

\begin{proof}
Following the proof for Theorem 1 in~\citep{dempster1977maximum}, from Eq.\ref{eq:log_likelihood_p_y_x_theta}, we have
\begin{equation}
    \begin{split}
         \log p_{\theta}(\mathbf{y}|\mathbf{x})&=\ell_{ELBO}(q,\theta)+KL[q(z)||p_{\theta}(z|\mathbf{y},\mathbf{x})], 
    \end{split}
\end{equation}
where $q(z)=p_{\theta^{old}}(z|\mathbf{y},\mathbf{x})$. Subtracting 
$\log p_{\theta}(\mathbf{y}|\mathbf{x})$ and $\log p_{\theta^{old}}(\mathbf{y}|\mathbf{x})$, we have
\begin{equation}
    \begin{split}
         \log  p_{\theta}(&\mathbf{y}|\mathbf{x}) - \log p_{\theta^{old}}(\mathbf{y}|\mathbf{x}) = \\ 
         & \ell_{ELBO}(q,\theta) - \ell_{ELBO}(q,\theta^{old}) + \\ 
         & KL[q(z)||p_{\theta}(z|\mathbf{y},\mathbf{x})] - KL[q(z)||p_{\theta^{old}}(z|\mathbf{y},\mathbf{x})].
    \end{split}
\end{equation}
Given that $KL[q(z)||p_{\theta}(z|\mathbf{y},\mathbf{x})] \ge KL[q(z)||p_{\theta^{old}}(z|\mathbf{y},\mathbf{x})]$ and that $\ell_{ELBO}(q,\theta) - \ell_{ELBO}(q,\theta^{old}) = 
        \mathbb{E}_{q(z)}[\log p_{\theta}(\mathbf{y},z|\mathbf{x})] - 
    \mathbb{E}_{q(z)}[\log p_{\theta^{old}}(\mathbf{y},z|\mathbf{x})]$, we conclude that
\begin{equation}
\begin{split}
    \log & p_{\theta}(\mathbf{y}|\mathbf{x}) - \log p_{\theta^{old}}  (\mathbf{y}|\mathbf{x})  \ge \\  
    & \mathbb{E}_{q(z)}[\log p_{\theta}(\mathbf{y},z|\mathbf{x})] - 
    \mathbb{E}_{q(z)}[\log p_{\theta^{old}}(\mathbf{y},z|\mathbf{x})] \ge 0
\end{split}    
    \label{eq:difference_bound}
\end{equation}

because of the assumption $\mathbb{E}_{q(z)}[\log p_{\theta}(\mathbf{y},z|\mathbf{x})] \ge 
    \mathbb{E}_{q(z)}[\log p_{\theta^{old}}(\mathbf{y},z|\mathbf{x})]$~\citep{dempster1977maximum}.

\end{proof}

\begin{restatable}[]{lemma}{Convergence}
    \label{thm:convergence}
    Suppose that $\{ \theta^{(e)}\}_{e=1}^{+\infty}$ denotes the sequence of trained model parameters from the maximisation of $\ell_{ELBO}$ in Eq.\ref{eq:M_step} such that:
    \begin{enumerate}
        \item the sequence $\{ \log p_{\theta^{(e)}}(\mathbf{y}|\mathbf{x})\}_{e=1}^{+\infty}$ is bounded above, and
        \item \footnotesize $\left(\mathbb{E}_{q(z)}[\log p_{\theta^{(e+1)}}(\mathbf{y},z|\mathbf{x})] - 
    \mathbb{E}_{q(z)}[\log p_{\theta^{(e)}}(\mathbf{y},z|\mathbf{x})]\right) \ge \xi \left(\theta^{(e+1)}-\theta^{(e)}\right)^{\top}\left(\theta^{(e+1)}-\theta^{(e)}\right)$, for $\xi>0$ and all $e \ge 1$, and $q(z)=p_{\theta^{(e)}}(z|\mathbf{y},\mathbf{x})$.  
    \end{enumerate}
    Then the sequence $\{\theta^{(e)}\}_{e=1}^{+\infty}$ converges to some $\theta^{\star} \in \Theta$.
 \end{restatable}

\begin{proof}
Following the proof for Theorem 2 in~\citep{dempster1977maximum}, the sequence $\{ \log p_{\theta^{(e)}}(\mathbf{y}|\mathbf{x})\}_{e=1}^{+\infty}$ is non-decreasing (from~\autoref{thm:correctness}) and bounded  above (from assumption (1) in~\autoref{thm:convergence}), so it converges to $L^{\star} < +\infty$. 
Therefore, according to Cauchy criterion~\citep{nguyen2020tutorial}, for any $\epsilon > 0$, we have $e^{(\epsilon)}$ such that, for $e \ge e^{(\epsilon)}$ and all $r \ge 1$,
\begin{equation}
\begin{split}
    \sum_{j=1}^{r} \big( \log p_{\theta^{(e+j)}}( & \mathbf{y}|\mathbf{x})  - 
    \log p_{\theta^{(e+j-1)}}(\mathbf{y}|\mathbf{x}) 
    \big) = \\
    & \big( \log p_{\theta^{(e+r)}}(\mathbf{y}|\mathbf{x}) - \log p_{\theta^{(e)}}(\mathbf{y}|\mathbf{x}) \big) < \epsilon.
\end{split}    
    \label{eq:bound_1_thm2}
\end{equation}
From~Eq.\ref{eq:difference_bound},
\begin{equation}
\begin{split}
    0 & \le  
    \mathbb{E}_{q(z)}[\log p_{\theta^{(e+j)}}(\mathbf{y},z|\mathbf{x})] - 
    \mathbb{E}_{q(z)}[\log p_{\theta^{(e+j-1)}}(\mathbf{y},z|\mathbf{x})] \\
    &\le
    \log p_{\theta^{(e+j)}}(\mathbf{y}|\mathbf{x}) - \log p_{\theta^{(e+j-1)}}(\mathbf{y}|\mathbf{x})
\end{split}    
    \label{eq:bound_2_thm2}
\end{equation}
for $j \ge 1$ and $q(z)=p_{\theta^{(e+j-1)}}(z|\mathbf{y},\mathbf{x})$. Hence, from Eq.\ref{eq:bound_1_thm2},
\begin{equation}
\begin{split}
    \sum_{j=1}^r \big( & \mathbb{E}_{q(z)}[\log p_{\theta^{(e+j)}}(\mathbf{y},z|\mathbf{x})] - \\
    & \mathbb{E}_{q(z)}[\log p_{\theta^{(e+j-1)}}(\mathbf{y},z|\mathbf{x})] \big) < \epsilon,
\end{split}    
    \label{eq:bound_3_thm2}
\end{equation}
for $e \ge e^{(\epsilon)}$ and all $r \ge 1$.
Given assumption (2) in ~\autoref{thm:convergence} for $e,e+1,e+2,...,e+r-1$, we
 have from Eq.\ref{eq:bound_3_thm2},
 \begin{equation}
     \epsilon > \xi \sum_{j=1}^{r} \left( \theta^{(e+j)} - \theta^{(e+j-1)} \right)^{\top}\left( \theta^{(e+j)} - \theta^{(e+j-1)} \right),
 \end{equation}
so
\begin{equation}
     \epsilon > \xi \left( \theta^{(e+r)} - \theta^{(e)} \right)^{\top}\left( \theta^{(e+r)} - \theta^{(e)} \right),
 \end{equation}
 which is a requirement to prove the convergence of $\theta^{(e)}$ to some $\theta^{\star} \in \Theta$.
\end{proof}

\section{Experiments} 
\label{sec:experiments}


\subsection{Experimental Setup}
\label{sec:experimental_setup}

We evaluate our method on CIFAR-10/-100~\citep{krizhevsky2009learning},  Controlled Noisy Web Labels (CNWL)~\citep{jiang2020beyond}, Clothing1M~\citep{xiao2015learning}, and  WebVision~\citep{li2017webvision}.
The CIFAR-10 and CIFAR-100 datasets contain 50,000 training images and 10,000 test images of size 32 $\times$ 32 pixels with 10 and 100 classes respectively. Since both these datasets have been annotated with clean labels, we use synthetic noise to evaluate the models. 
For CIFAR-10/-100, we evaluate three types of noise: symmetric~\citep{tanaka2018joint,li2019learning}, asymmetric~\citep{tanaka2018joint,li2019learning}, and semantic~\citep{rog}.
For symmetric noise we used $\eta \in \{0.2, 0.5, 0.8, 0.9\}$, where $\eta$ was defined in Section Method as the symmetric noise probability. 
The asymmetric noise was applied to the dataset, similarly to~\citep{li2020dividemix}, which replaces the labels \emph{truck} $\rightarrow$ \emph{automobile}, \emph{bird} $\rightarrow$ \emph{airplane}, \emph{deer} $\rightarrow$ \emph{horse}, and \emph{cat} $\rightarrow$ \emph{dog}. For asymmetric noise, we use the noise rates of 40\% and 49\%. For the semantic noise, we use the same setup from~\citep{rog}, which generates semantically noisy labels based on a trained VGG~\citep{vgg}, DenseNet~(DN)~\citep{huang2017densely}, and ResNet~(RN)~\citep{ResNet18} on CIFAR-10 and CIFAR-100.

The CNWL dataset~\citep{jiang2020beyond} is a benchmark to study real-world web label noise in a controlled setting. Both images and labels are crawled from the web and the noisy labels are determined by matching images. The controlled setting provide different magnitudes of label corruption in real applications, varying from 0 to 80\%. CNWL provides controlled web noise for Mini-ImageNet dataset, called red noise. The red Mini-ImageNet consists of 50k training images and 5000 test images, with 100 classes. The original image sizes are of 84$\times$84 pixels, which are resized to 32$\times$32 pixels. The noise rates use in this work are 20\%, 40\%, 60\% and 80\%, as used in~\citep{FaMUS}.   

Clothing1M is a dataset of 14 classes containing 1 million training images downloaded from online shopping websites.
All training images are resized to $256 \times 256$ pixels~\citep{li2020dividemix, han2019deep}. 
The noise rate is estimated to be asymmetric~\citep{yi2019probabilistic} with a rate of 40\%~\citep{xiao2015learning} and class distribution is heavily imbalanced.
Clothing1M has 50k and 14k clean images for 
training and validation, respectively, but we do not use them for training. 
The testing set has 10k clean-labelled images.

The WebVision~\citep{li2017webvision} is a real-world large scale dataset containing 2.4 million images collected from the  internet,  with  the  same 1000 classes from ImageNet~\citep{deng2009imagenet}. As the images vary in size, we resized them to  $227 \times 227$ pixels. WebVision provides a clean test set of 50k images, with 50 images per class.  We compare our model using the first 50 classes of the Google image subset, as in \citep{li2020dividemix, chen2019understanding}

All experiments were run on Intel Core i9 computer with 128GB memory and 4x nVidia GeForce RTX 3090.

\begin{table}
\centering
\scalebox{0.8}{
\begin{tabular}{lcccc}
\toprule
Method/ noise ratio & 20\% & 40\% & 60\%& 80\% \\
\midrule
 Cross-entropy~\citep{FaMUS}    & 47.36 & 42.70 & 37.30 & 29.76 \\
 Mixup~\citep{zhang2017mixup}          & 49.10 & 46.40 & 40.58 & 33.58 \\
 DivideMix~\citep{li2020dividemix}     & 50.96 & 46.72 & 43.14 & 34.50 \\
 MentorMix~\citep{jiang2020beyond}  & 51.02 & 47.14 & 43.80 & 33.46 \\
 FaMUS~\citep{FaMUS}  & 51.42 & 48.06 & 45.10 & 35.50 \\
 \textbf{ScanMix (Ours)} & \textbf{59.06} & \textbf{54.54} & \textbf{52.36} & \textbf{40.00} \\
\bottomrule \\
\end{tabular}
}
\caption{Test accuracy (\%) for Red Mini-ImageNet. Results from baseline methods are as presented in \citep{FaMUS}. Top methods within $1\%$ are in \textbf{bold}.}
\label{tab:results_red_noise}
\end{table}

\begin{table}[t!]
\centering
\scalebox{0.8}{
\begin{tabular}{lcccc}
\toprule
 dataset & \multicolumn{2}{c}{WebVision} &  \multicolumn{2}{c}{ILSVRC12} \\
\midrule
{Method} & {Top-1} & {Top-5} & {Top-1} & {Top-5} \\
\midrule
 F-correction~\citep{patrini2017making}    & 61.12 & 82.68 & 57.36 & 82.36 \\
 Decoupling~\citep{malach2017decoupling}     & 62.54 & 84.74 & 58.26 & 82.26\\
 D2L~\citep{ma2018dimensionality}            & 62.68 & 84.00  & 57.80 & 81.36\\
 MentorNet~\citep{jiang2018mentornet}      & 63.00  & 81.40 & 57.80 & 79.92 \\
 Co-teaching~\citep{han2018co}    & 63.58 & 85.20 & 61.48 & 84.70 \\
 Iterative-CV~\citep{chen2019understanding}   & 65.24 & 85.34 & 61.60 & 84.98\\
 MentorMix~\citep{jiang2020beyond}& 76.00 & 90.20 &  72.90 & 91.10 \\

 DivideMix~\citep{li2020dividemix}      & 77.32  & 91.64 & \textbf{75.20} & 90.84\\
 ELR+~\citep{liu2020early}& 77.78 & 91.68 & 70.29 & 89.76 \\

  MOIT+~\citep{ortego2020multi}& 78.76 & - & - & - \\
  FSR~\citep{zhang2021FSR}& 74.90 & 88.20 & 72.30 & 87.20 \\
 \textbf{ScanMix~(Ours)}     & \textbf{80.04} & \textbf{93.04} & \textbf{75.76} & \textbf{92.60} \\
 \bottomrule \\
\end{tabular}
}
\caption{Test accuracy (\%) for WebVision~\citep{li2017webvision} by methods trained with 100 epochs. Baseline results are as presented in \citep{li2020dividemix}. Top methods within $1\%$ are in \textbf{bold}.}

\label{tab:res_WebVision}
\end{table}

\begin{table}[t]
\footnotesize
\centering
\scalebox{1.0}{
\begin{tabular}{lcc}
\toprule
Method & Test Accuracy \\
\midrule
 Cross-Entropy~\citep{li2020dividemix}  & 69.21  \\

 M-correction \citep{arazo2019unsupervised}   & 71.00 \\

  Meta-Cleaner~\citep{zhang2019metacleaner}    & 72.50  \\
  Meta-Learning~\citep{li2019learning}   & 73.47 \\
 PENCIL\citep{yi2019probabilistic} &   73.49 \\

 DeepSelf~\citep{han2019deep} & \textbf{74.45} \\

 CleanNet~\citep{lee2018cleannet}    & \textbf{74.69} \\

 DivideMix~\citep{li2020dividemix}      & \textbf{74.76} \\

ScanMix (Ours) & \textbf{74.35} \\
 \bottomrule
 \\
\end{tabular}
}
\caption{Results on Clothing1M~\citep{xiao2015learning} for ScanMix and SOTA approaches (SOTA results collected from \citep{li2020dividemix} or original papers). Top results within 1\% are highlighted in \textbf{bold}.
}
\label{tab:sota_cl}
\end{table}

\begin{table*}[t!]
\centering
\scalebox{0.65}{
\begin{tabular}{lc|cccc|cc||cccc}
\toprule
\multicolumn{2}{c}{dataset} & \multicolumn{6}{c}{CIFAR-10} & \multicolumn{4}{c}{CIFAR-100}\\    
\midrule
\multicolumn{2}{c}{Noise type} & \multicolumn{4}{c}{sym.} & \multicolumn{2}{c}{asym.} &  \multicolumn{4}{c}{sym.} \\
\midrule

Method/ noise ratio & &  20\% & 50\% & 80\% & 90\% & 40\%& 49\% & 20\% & 50\% & 80\% & 90\% \\
\midrule

\multirow{1}{*}{Self-superv. pre-train =(SCAN)~\citep{SCAN}}& & 81.6 & 81.6 & 81.6 & 81.6 & 81.6 & 81.6 & 44.0 & 44.0 & 44.0 & 44.0 \\

\multirow{1}{*}{Self-superv. pre-train* (SCAN)~\citep{SCAN}}& & 77.5 & 77.5 & 77.5 & 77.5 & 77.5 & 77.5 & 37.1 & 37.1 & 37.1 & 37.1\\

\multirow{1}{*}{SSL (DivideMix)~\citep{li2020dividemix}}& & \textbf{96.1} & \textbf{94.6} & \textbf{93.2} & 76.0& \textbf{93.4} & 83.7* & \textbf{77.3} & 74.6 & 60.2 & 31.5 \\

\multirow{1}{*}{Self-superv. pre-train + SSL (DivideMix)*}& & 
\textbf{95.3} &
\textbf{94.4} &
\textbf{93.7} &
\textbf{91.0} &
\textbf{93.3} &
85.9 &
75.2 &
74.4 &
64.4 &
52.8 \\

\multirow{1}{*}{\textbf{ScanMix (Ours)}}& & \textbf{96.0} & \textbf{94.5} & \textbf{93.5} & \textbf{91.0} & \textbf{93.7} & \textbf{88.6} & \textbf{77.0} & \textbf{75.7} & \textbf{66.0} & \textbf{58.5}\\
\bottomrule \\
\end{tabular}
}
\caption{In this ablation study we show the classification accuracy in the testing set of CIFAR-10 and CIFAR-100 under symmetric and asymmetric noises at several rates.  First, we show the results of self-supervised pre-training using the current SOTA SCAN~\citep{SCAN} (first two rows, with the results with (*) produced by locally running the published code provided by the authors).  Then we show the current SOTA SSL learning for noisy label DivideMix~\citep{li2020dividemix}.  Next, we show the results of DivideMix pre-trained with SCAN.  The last row shows our ScanMix that combines SSL and semantic clustering. The top results within $1\%$ are highlighted in \textbf{bold}.}
\label{tab:results_ablation}
\end{table*}

\subsection{Implementation}
\label{sec:implementation}

\paragraph{CIFAR-10/-100}
We use PreAct-ResNet-18 
as our backbone model following~\citep{li2020dividemix}. For the self-supervised pre-training learning task, we adopt the standard SimCLR~\citep{SimCLR} implementation with a batch size of 512, SGD optimiser with a learning rate of 0.4, decay rate of 0.1, momentum of 0.9 and weight decay of 0.0001, and run it for 500 epochs.  This pre-trained model produces feature representations of 128 dimensions. Using these representations we mine $K = 20$ nearest neighbours (as in \citep{SCAN}) for each sample to form the sets $\{ \mathcal{N}_{\mathbf{x}_{i}}\}_{i=1}^{|\mathcal{D}|}$, defined in the Method Section. 
For the semantic clustering task, we use a batch size of 128, $\lambda_e = 2$ as in \citep{SCAN}, SGD optimiser with momentum of 0.9, weight decay of 0.0005 and learning rate $\in \{0.001, 0.00001\}$ based on the predicted noise rate, which is estimated with $|\mathcal{U}|/|\mathcal{D}|$, defined in~Eq.\ref{eq:clean_noisy_sets} -- if this ratio is larger than $0.6$, then the learning rate is $0.001$, otherwise, the learning rate is $0.00001$.  This accounts for the fact that when the estimated label noise is high, then we want to increase the influence of semantic clustering in the training; but when the label noise is low, then the signal from the labels in the SSL method should carry more weighting. For the SSL, we adopt the implementation of~\citep{li2020dividemix} and use the same hyperparameters, where we rely on SGD with learning rate of 0.02 (which is reduced to 0.002 halfway through the training), momentum of 0.9 and weight decay of 0.0005. Number of epochs $E$ = 300. 

\paragraph{Red Mini-ImageNet} We use PreAct-ResNet-18 as our backbone model, following~\citep{FaMUS}. For the self-supervised pre-training, we adopt the standard SimCLR~\citep{SimCLR} implementation with batch size 128. All other parameters for the self-supervised pre-training and semantic clustering are the same as for CIFAR, except for the semantic clustering learning rate, which we used 0.001, and the $\lambda_u=0$ for all noise rates. The feature representation learned from this process has 128 dimensions. For the SSL, we adopt the implementation of~\citep{FaMUS}, where we train for 300 epochs,  relying on SGD with learning rate of 0.02 (decreased by a factor of ten at epoch 200 and epoch 250), momentum of 0.9 and weight decay of 5e-4. We also resized the images from $84\times84$ to $32\times32$~\citep{FaMUS}.

\paragraph{Clothing1M} We use ResNet-50 as our backbone model, which is trained for 80 epochs with a WarmUp stage of 1 epoch. For the self-supervised pre-training task we adopt the standard MoCo-v2 method for a 4-GPU training~\citep{MoCoV2} with a batch size of 128, SGD optimiser with a learning rate of 0.015, momentum of 0.9 and weight decay of 0.0001 and run it for 100 epochs. In this pre-training task we use 100k randomly selected images from Clothing1M training set as the pre-training images. All the other parameters were the same as described above for CIFAR, except for the batch size of semantic clustering task was 64 and the number of epochs $E$=80. During ScanMix training, we followed~\citep{li2020dividemix}, which relies on 64k randomly selected training images from the entire training for each epoch. As the training images change for every epoch, we adapted ScanMix to update the nearest neighbors before training each batch. Different from ~\citep{li2020dividemix}, we do not use the pre-trained weights from ImageNet.

\paragraph{WebVision}
We use InceptionResNet-V2 as our backbone model, following~\citep{li2020dividemix}. For the self-supervised pre-training task we adopt the standard MoCo-v2 method for a 4-GPU training~\citep{MoCoV2} with a batch size of 128, SGD optimiser with a learning rate of 0.015, momentum of 0.9 and weight decay of 0.0001, and run it for 100 epochs with a WarmUp stage of 1 epoch. The feature representations learned from this process have 128 dimensions. All the other parameters were the same as described above for CIFAR, except the batch size of semantic clustering task was 64, and number of epochs $E$ = 100.

\subsection{Comparison with State-of-the-Art}
\label{sec:comparison_SOTA}

We compare ScanMix with several existing methods using the datasets described in Sec. Experimental Setup. For CIFAR-10 and CIFAR-100 in Table~\ref{tab:results_cifar}, we evaluate the models using different  levels  of  symmetric  label  noise, ranging from  20\%  to  90\% and asymmetric noise rates of 40\% and 49\%.  We report both the best test accuracy across all epochs and the averaged test accuracy over the last 10 epochs of training.  

Results show that our method significantly outperforms the previous methods under severe label noise. Specifically, we observe an increase of roughly +13\% for CIFAR-10 with 90\% symmetric noise, +5\% for CIFAR-10 with 49\% asymmetric noise, +25\% for CIFAR-100 with 90\% symmetric noise and +5\% for CIFAR-100 with 80\% symmetric noise. These results show that our ScanMix does make the model more robust to noisy labels than previous methods, particularly for severe label noise. To demonstrate that more clearly, we computed mean and variance accuracy using bootstrapping, and applied a T-test to compare ScanMix and DivideMix. For all cases that we claim to be better in Table~\ref{tab:results_cifar} (symmetric at 80\% and 90\% on Cifar10,100 and asymmetric at 40\% and 49\% on Cifar10), we obtained p-values $<0.01$. 

Table~\ref{tab:res_semantic} shows the ability of our method to handle semantic noise, which can be regarded as a harder and more realistic type of label noise that depends not only on label transition, but also on the image features.  

The current SOTA for this benchmark is RoG~\citep{rog}, and even though the noise rates are not particularly large, our ScanMix shows results that are better by a large margin varying from 12\% to 22\%.
The results on Red Mini-ImageNet~\citep{FaMUS} in~\autoref{tab:results_red_noise} shows that ScanMix provides substantial gains form 4\% to 7\% over the SOTA for all noise rates. 

We also evaluate ScanMix on the noisy large-scale dataset WebVision. Table \ref{tab:res_WebVision} shows the Top-1/-5 test accuracy using the WebVision and ILSVRC12 test sets. Results show that ScanMix is slightly better than the SOTA for both WebVision test sets and top-5 ILSVRC12 test set. This suggests that our approach is also effective in large-scale, low noise rate problems.
Results on Clothing1M in Tab.~\ref{tab:sota_cl} show that ScanMix is on par with the current SOTA in the field, even though our method does use the whole training set for the pre-training stage (recall that we randomly selected 100k out of the 1M training images for pre-training) and differently from most of previous approaches, we do not rely on an ImageNet pre-trained model, as explained in Sec.~\ref{sec:implementation}. These two issues should have had a significant negative impact on the performance of ScanMix, but these Clothing1M results indicate that ScanMix remained robust in this challenging scenario. 

For the running time complexity, ScanMix and DivideMix are similar asymptotically since both have linear complexity in terms of the training set size, as described in the Section~\ref{sec:scanmix}. In practice, ScanMix is two times slower. On CIFAR-10, DivideMix takes 13.93 GPU hours while ScanMix takes 27.94 hours (where pre-train takes 5.9 GPU hours).

\subsection{Ablation Study}

We show the results of the ablation study of ScanMix in Table~\ref{tab:results_ablation}.  Using classification accuracy in the testing set of CIFAR-10 and CIFAR-100 under symmetric and asymmetric noises at several rates, we aim to show the influence of self-supervised training by itself or in combination with SSL.  For self-supervised learning, we use the current SOTA method, SCAN~\citep{SCAN}, displayed in the first two rows, with the first row containing the published results, and the second, our replicated results using the authors' code.  The result is the same across different noise rates because it never uses the noisy labels for training.
Using the pure SSL method, DivideMix~\citep{li2020dividemix}, which is the current SOTA in noisy label learning, we see that it has much better results for low noise levels, but SCAN is better for severe label noise. 
When using SCAN for pre-training DivideMix, we note that results become quite good for all types and levels of noise.  Neverthless, our ScanMix improves the results of SCAN + DivideMix, showing the efficacy of ScanMix, which combines SSL with semantic clustering.

\begin{figure}
\centering
\begin{tabular}{cc}
  \includegraphics[width=0.45\linewidth]{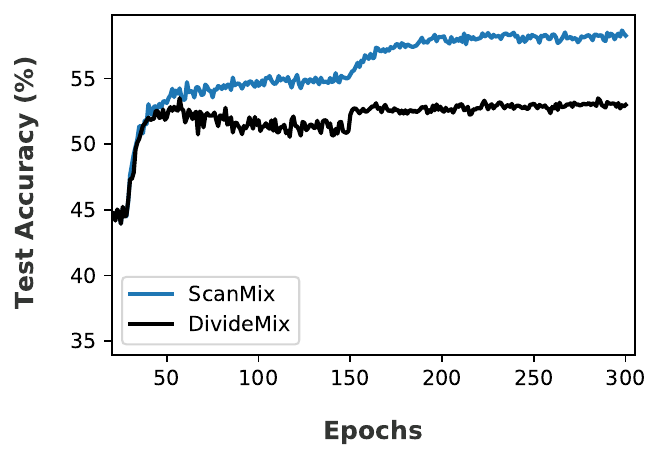}   &  
  \includegraphics[width=0.45\linewidth]{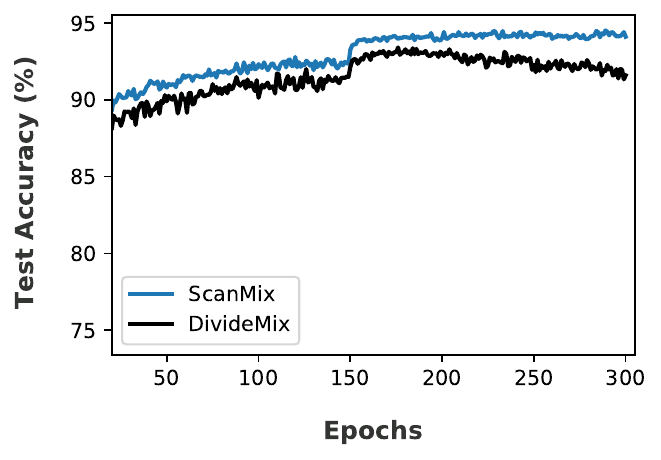} \\
  a) 90\% symmetric on CIFAR-100  & b) 40\% asymmetric on CIFAR-10
\end{tabular}
  \caption{Test accuracy (\%) as a function of the number of training epochs for ScanMix (blue) and DivideMix (black) for 90\% asymmetric noise on CIFAR-100 (a), and 40\% asymmetric noise on CIFAR-10 (b).}
  \label{fig:scanmix_loss_testing_epoch}
\end{figure}

A common issue with learning with noisy labels is the tendency of models to overfit the noisy labels during training~\citep{liu2020early,ma2018dimensionality}, causing a reduction of accuracy in the test set. To show the robustness of ScanMix to this issue, we present in Fig.~\ref{fig:scanmix_loss_testing_epoch} the prediction accuracy on the test set as a function of the number of training epochs for ScanMix (blue) and DivideMix (black) for 90\% asymmetric noise on CIFAR-100 (a), and 40\% asymmetric noise on CIFAR-10 (b). Notice that in both cases, ScanMix is shown to be more robust to overfitting than DivideMix.

\begin{figure}
\centering
  \includegraphics[width=0.7\linewidth]{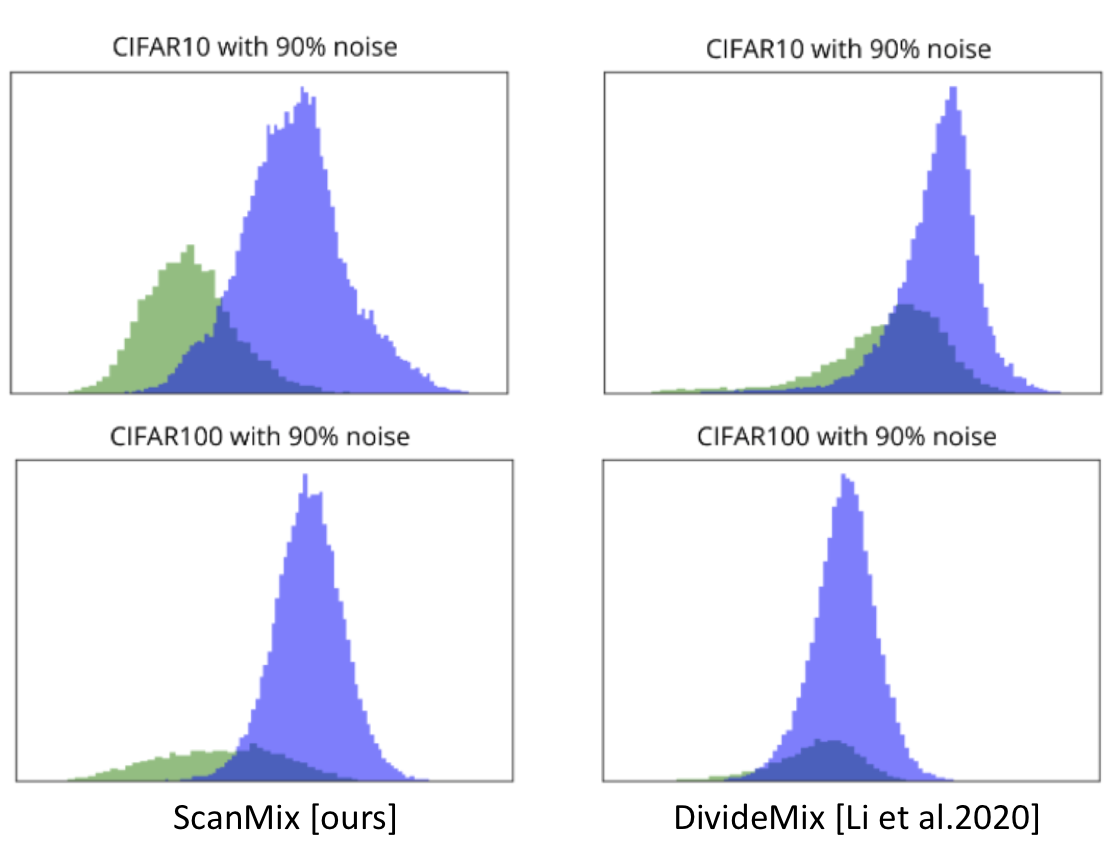}
  \caption{Per-sample normalised loss distributions of the training set produced in the early stages of the training by our ScanMix (left) and DivideMix~\citep{li2020dividemix} (right) for CIFAR-10 (top) and CIFAR-100 (bottom) affected by 90\% label noise, where green bars represent the clean samples and blue bars the noisy samples.}
  \label{fig:scanmix_vs_li2020dividemix}
\end{figure}

We also demonstrate that ScanMix is able to provide a reliable separation between clean and noisy samples.
Figure~\ref{fig:scanmix_vs_li2020dividemix} shows a comparison between the distributions of losses produced by ScanMix and DivideMix at one of the early epochs of training for CIFAR-10 and CIFAR-100 affected by 90\% label noise.  This figure shows that semantic clustering combined with SSL in ScanMix enables a much clearer separation between the clean (green bars) and noisy (blue bars) samples, when compared with the distribution produced by DivideMix.  Such clearer separation will help the classification of clean samples in~Eq.\ref{eq:clean_noisy_sets}, which in turn will improve the performance of the SSL in~Eq.\ref{eq:max_likelihood_optimisation}.

\section{Conclusion and Future Work}
\label{sec:conclusions}

In this work we presented ScanMix, a novel training strategy that produces superior robustness to severe label noise and competitive robustness to non-severe label noise problems, compared with the SOTA.
Results on CIFAR-10/-100, Red  Mini-ImageNet, Clothing1M and WebVision showed that our proposed ScanMix outperformed SOTA methods, with large improvements particularly in severe label noise problems.
Our approach also produced superior results for semantic noise and real-world web label noise, which are regarded to be the most challenging noise types. 
These results show evidence for our claims in Section~\ref{sec:introduction}, that SSL noisy label learning methods (e.g., DivideMix~\citep{li2020dividemix}) depend on an effective way to classify clean and noisy samples, which works well for small noise rates, but not for severe noise rates.  Semantic clustering methods (e.g., SCAN~\citep{SCAN}) ignore labels, enabling them to work well for severe noise rates, but poorly for low noise.  
Hence, our ScanMix explores the advantages of SSL and semantic clustering to achieve SOTA results for severe label noise rates, while being competitive for non-severe label noise.

The increasing availability of large-scale datasets is associated with a decreasing availability of trustworthy annotations. This can introduce label noise into training sets, and reduce the generalisation ability of machine learning models.
Our method can mitigate this issue and enable the use of large-scale datasets by communities that do not have other ways to re-annotate such datasets, thus democratising machine learning.
A drawback of our approach is the longer training time, compared with the SOTA DivideMix~\citep{li2020dividemix}, so we are currently working on an approach that mitigates this issue by having a joint self-supervised and semi-supervised training algorithm.
Another point that can be improved in ScanMix is the semantic clustering algorithm, which can explore more robust methods, such as RBSMF~\citep{wang2020robust}, MPF~\citep{wang2018detecting}, ClusterNet~\citep{shukla2020semi}, and USADTM~\citep{han2020unsupervised}.

\section*{Acknowledgments}

This work was supported by Australian Research Council through grant FT190100525.  


\bibliography{ScanMix}

\begin{thebibliography}{10}
\expandafter\ifx\csname url\endcsname\relax
  \def\url#1{\texttt{#1}}\fi
\expandafter\ifx\csname urlprefix\endcsname\relax\def\urlprefix{URL }\fi
\expandafter\ifx\csname href\endcsname\relax
  \def\href#1#2{#2} \def\path#1{#1}\fi

\bibitem{litjens2017survey}
G.~Litjens, T.~Kooi, B.~E. Bejnordi, A.~A.~A. Setio, F.~Ciompi, M.~Ghafoorian,
  J.~A. Van Der~Laak, B.~Van~Ginneken, C.~I. S{\'a}nchez, A survey on deep
  learning in medical image analysis, Medical image analysis 42 (2017) 60--88.

\bibitem{frenay_survey}
B.~Fr{\'e}nay, M.~Verleysen, Classification in the presence of label noise: a
  survey, IEEE transactions on neural networks and learning systems 25~(5)
  (2013) 845--869.

\bibitem{Zhang2017UnderstandingDL}
C.~Zhang, S.~Bengio, M.~Hardt, B.~Recht, O.~Vinyals, Understanding deep
  learning requires rethinking generalization, in: International Conference on
  Learning Representations (ICLR), 2017.

\bibitem{ding2018semi}
Y.~Ding, L.~Wang, D.~Fan, B.~Gong, A semi-supervised two-stage approach to
  learning from noisy labels, in: 2018 IEEE Winter Conference on Applications
  of Computer Vision (WACV), IEEE, 2018, pp. 1215--1224.

\bibitem{ortego2019towards}
D.~Ortego, E.~Arazo, P.~Albert, N.~E. O'Connor, K.~McGuinness, Towards robust
  learning with different label noise distributions, in: 2020 25th
  International Conference on Pattern Recognition (ICPR), IEEE, 2021, pp.
  7020--7027.

\bibitem{ortego2020multi}
D.~Ortego, E.~Arazo, P.~Albert, N.~E. O'Connor, K.~McGuinness, Multi-objective
  interpolation training for robustness to label noise, in: Conference on
  Computer Vision and Pattern Recognition (CVPR), 2021.

\bibitem{li2020dividemix}
J.~Li, R.~Socher, S.~C. Hoi, Dividemix: Learning with noisy labels as
  semi-supervised learning, in: International Conference on Learning
  Representations (ICLR), 2020.

\bibitem{MoCo}
K.~He, H.~Fan, Y.~Wu, S.~Xie, R.~Girshick, Momentum contrast for unsupervised
  visual representation learning, in: Proceedings of the IEEE/CVF Conference on
  Computer Vision and Pattern Recognition, 2020, pp. 9729--9738.

\bibitem{MoCoV2}
X.~{Chen}, H.~{Fan}, R.~{Girshick}, K.~{He}, {Improved Baselines with Momentum
  Contrastive Learning}, arXiv e-prints (2020) arXiv:2003.04297\href
  {http://arxiv.org/abs/2003.04297} {\path{arXiv:2003.04297}}.

\bibitem{SCAN}
W.~Van~Gansbeke, S.~Vandenhende, S.~Georgoulis, M.~Proesmans, L.~Van~Gool,
  Scan: Learning to classify images without labels, in: European Conference on
  Computer Vision, Springer, 2020, pp. 268--285.

\bibitem{SimCLR}
T.~Chen, S.~Kornblith, M.~Norouzi, G.~Hinton, A simple framework for
  contrastive learning of visual representations, in: International Conference
  on Machine Learning (ICML), PMLR, 2020, pp. 1597--1607.

\bibitem{SELF}
T.~Nguyen, C.~Mummadi, T.~Ngo, L.~Beggel, T.~Brox, Self: learning to filter
  noisy labels with self-ensembling, in: International Conference on Learning
  Representations (ICLR), 2020.

\bibitem{dempster1977maximum}
A.~P. Dempster, N.~M. Laird, D.~B. Rubin, Maximum likelihood from incomplete
  data via the em algorithm, Journal of the Royal Statistical Society: Series B
  (Methodological) 39~(1) (1977) 1--22.

\bibitem{rebbapragada2007class}
U.~Rebbapragada, C.~E. Brodley, Class noise mitigation through instance
  weighting, in: European conference on machine learning, Springer, 2007, pp.
  708--715.

\bibitem{krizhevsky2009learning}
A.~Krizhevsky, G.~Hinton, et~al., Learning multiple layers of features from
  tiny images, Citeseer, 2009.

\bibitem{FaMUS}
Y.~Xu, L.~Zhu, L.~Jiang, Y.~Yang, Faster meta update strategy for noise-robust
  deep learning, in: Conference on Computer Vision and Pattern Recognition
  (CVPR), 2021.

\bibitem{xiao2015learning}
T.~Xiao, T.~Xia, Y.~Yang, C.~Huang, X.~Wang, Learning from massive noisy
  labeled data for image classification, in: Proceedings of the IEEE conference
  on computer vision and pattern recognition, 2015, pp. 2691--2699.

\bibitem{li2017webvision}
W.~Li, L.~Wang, W.~Li, E.~Agustsson, L.~Van~Gool, Webvision database: Visual
  learning and understanding from web data, arXiv preprint arXiv:1708.02862,
  2017.

\bibitem{jaehwan2019photometric}
L.~Jaehwan, Y.~Donggeun, K.~Hyo-Eun, Photometric transformer networks and label
  adjustment for breast density prediction, in: Proceedings of the IEEE
  International Conference on Computer Vision Workshops, 2019.

\bibitem{yuan2018iterative}
B.~Yuan, J.~Chen, W.~Zhang, H.-S. Tai, S.~McMains, Iterative cross learning on
  noisy labels, in: 2018 IEEE Winter Conference on Applications of Computer
  Vision (WACV), IEEE, 2018, pp. 757--765.

\bibitem{zhang2021learning}
Y.~Zhang, S.~Zheng, P.~Wu, M.~Goswami, C.~Chen, Learning with feature dependent
  label noise: a progressive approach, in: International Conference on Feature
  Representation (ICLR), 2021.

\bibitem{liu2020early}
S.~Liu, J.~Niles-Weed, N.~Razavian, C.~Fernandez-Granda, Early-learning
  regularization prevents memorization of noisy labels, in: Conference on
  Neural Information Processing Systems (NeurIPS), 2020.

\bibitem{wang2019imae}
X.~Wang, Y.~Hua, E.~Kodirov, N.~M. Robertson, Imae for noise-robust learning:
  Mean absolute error does not treat examples equally and gradient magnitude's
  variance matters, arXiv preprint arXiv:1903.12141, 2019.

\bibitem{wang2019symmetric}
Y.~Wang, X.~Ma, Z.~Chen, Y.~Luo, J.~Yi, J.~Bailey, Symmetric cross entropy for
  robust learning with noisy labels, in: Proceedings of the IEEE International
  Conference on Computer Vision, 2019, pp. 322--330.

\bibitem{han2018pumpout}
B.~Han, G.~Niu, J.~Yao, X.~Yu, M.~Xu, I.~Tsang, M.~Sugiyama, Pumpout: A meta
  approach for robustly training deep neural networks with noisy labels, in:
  ICML Workshop on Uncertainty and Robustness in Deep Learning, 2019.

\bibitem{sun2021learning}
H.~Sun, C.~Guo, Q.~Wei, Z.~Han, Y.~Yin, Learning to rectify for robust learning
  with noisy labels, Pattern Recognition 124 (2022) 108467.

\bibitem{ren2018learning}
M.~Ren, W.~Zeng, B.~Yang, R.~Urtasun, Learning to reweight examples for robust
  deep learning, in: International conference on machine learning, PMLR, 2018,
  pp. 4334--4343.

\bibitem{miao2015rboost}
Q.~Miao, Y.~Cao, G.~Xia, M.~Gong, J.~Liu, J.~Song, Rboost: Label noise-robust
  boosting algorithm based on a nonconvex loss function and the numerically
  stable base learners, IEEE transactions on neural networks and learning
  systems 27~(11) (2015) 2216--2228.

\bibitem{tarvainen2017mean}
A.~Tarvainen, H.~Valpola, Mean teachers are better role models: Weight-averaged
  consistency targets improve semi-supervised deep learning results, in:
  Advances in neural information processing systems, 2017, pp. 1195--1204.

\bibitem{jiang2018mentornet}
L.~Jiang, Z.~Zhou, T.~Leung, L.-J. Li, L.~Fei-Fei, Mentornet: Learning
  data-driven curriculum for very deep neural networks on corrupted labels, in:
  International Conference on Machine Learning, 2018, pp. 2304--2313.

\bibitem{malach2017decoupling}
E.~Malach, S.~Shalev-Shwartz, Decoupling" when to update" from" how to update",
  in: Advances in Neural Information Processing Systems, 2017, pp. 960--970.

\bibitem{han2018co}
B.~Han, Q.~Yao, X.~Yu, G.~Niu, M.~Xu, W.~Hu, I.~Tsang, M.~Sugiyama,
  Co-teaching: Robust training of deep neural networks with extremely noisy
  labels, in: Advances in neural information processing systems, 2018, pp.
  8527--8537.

\bibitem{yu2019does}
X.~Yu, B.~Han, J.~Yao, G.~Niu, I.~Tsang, M.~Sugiyama, How does disagreement
  help generalization against label corruption?, in: International Conference
  on Machine Learning, PMLR, 2019, pp. 7164--7173.

\bibitem{ma2018dimensionality}
X.~Ma, Y.~Wang, M.~E. Houle, S.~Zhou, S.~Erfani, S.~Xia, S.~Wijewickrema,
  J.~Bailey, Dimensionality-driven learning with noisy labels, in:
  International Conference on Machine Learning, 2018, pp. 3355--3364.

\bibitem{yu2018learning}
X.~Yu, T.~Liu, M.~Gong, D.~Tao, Learning with biased complementary labels, in:
  Proceedings of the European Conference on Computer Vision (ECCV), 2018, pp.
  68--83.

\bibitem{kim2019nlnl}
Y.~Kim, J.~Yim, J.~Yun, J.~Kim, Nlnl: Negative learning for noisy labels, in:
  Proceedings of the IEEE International Conference on Computer Vision, 2019,
  pp. 101--110.

\bibitem{zhang2019metacleaner}
W.~Zhang, Y.~Wang, Y.~Qiao, Metacleaner: Learning to hallucinate clean
  representations for noisy-labeled visual recognition, in: Proceedings of the
  IEEE Conference on Computer Vision and Pattern Recognition, 2019, pp.
  7373--7382.

\bibitem{jiang2020beyond}
L.~Jiang, D.~Huang, M.~Liu, W.~Yang, Beyond synthetic noise: Deep learning on
  controlled noisy labels, International Conference on Machine Learning (ICML),
  2020.

\bibitem{zhang2020distilling}
Z.~Zhang, H.~Zhang, S.~O. Arik, H.~Lee, T.~Pfister, Distilling effective
  supervision from severe label noise, in: Proceedings of the IEEE/CVF
  Conference on Computer Vision and Pattern Recognition, 2020, pp. 9294--9303.

\bibitem{sachdeva2021evidentialmix}
R.~Sachdeva, F.~R. Cordeiro, V.~Belagiannis, I.~Reid, G.~Carneiro,
  Evidentialmix: Learning with combined open-set and closed-set noisy labels,
  in: Proceedings of the IEEE/CVF Winter Conference on Applications of Computer
  Vision, 2021, pp. 3607--3615.

\bibitem{cordeiro2021longremix}
F.~R. Cordeiro, R.~Sachdeva, V.~Belagiannis, I.~Reid, G.~Carneiro, Longremix:
  Robust learning with high confidence samples in a noisy label environment,
  arXiv preprint arXiv:2103.04173.

\bibitem{MixMatch}
D.~Berthelot, N.~Carlini, I.~Goodfellow, N.~Papernot, A.~Oliver, C.~Raffel,
  Mixmatch: A holistic approach to semi-supervised learning, in: Neural
  Information Processing Systems (NeurIPS), 2019.

\bibitem{zhang2017mixup}
H.~Zhang, M.~Cisse, Y.~N. Dauphin, D.~Lopez-Paz, mixup: Beyond empirical risk
  minimization, in: International Conference on Learning Representations
  (ICLR), 2018.

\bibitem{wang2018detecting}
Q.~Wang, M.~Chen, F.~Nie, X.~Li, Detecting coherent groups in crowd scenes by
  multiview clustering, IEEE transactions on pattern analysis and machine
  intelligence 42~(1) (2018) 46--58.

\bibitem{wang2020robust}
Q.~Wang, X.~He, X.~Jiang, X.~Li, Robust bi-stochastic graph regularized matrix
  factorization for data clustering, IEEE Transactions on Pattern Analysis and
  Machine Intelligence 44~(1) (2020) 390--403.

\bibitem{shukla2020semi}
A.~Shukla, G.~S. Cheema, S.~Anand, Semi-supervised clustering with neural
  networks, in: 2020 IEEE Sixth International Conference on Multimedia Big Data
  (BigMM), IEEE, 2020, pp. 152--161.

\bibitem{han2020unsupervised}
T.~Han, J.~Gao, Y.~Yuan, Q.~Wang, Unsupervised semantic aggregation and
  deformable template matching for semi-supervised learning, Advances in Neural
  Information Processing Systems 33 (2020) 9972--9982.

\bibitem{chiaroni2019hallucinating}
F.~Chiaroni, M.-C. Rahal, N.~Hueber, F.~Dufaux, Hallucinating a cleanly labeled
  augmented dataset from a noisy labeled dataset using gan, in: 2019 IEEE
  International Conference on Image Processing (ICIP), IEEE, 2019, pp.
  3616--3620.

\bibitem{patrini2017making}
G.~Patrini, A.~Rozza, A.~Krishna~Menon, R.~Nock, L.~Qu, Making deep neural
  networks robust to label noise: A loss correction approach, in: Proceedings
  of the IEEE Conference on Computer Vision and Pattern Recognition, 2017, pp.
  1944--1952.

\bibitem{rog}
K.~Lee, S.~Yun, K.~Lee, H.~Lee, B.~Li, J.~Shin, Robust inference via generative
  classifiers for handling noisy labels, in: International Conference on
  Machine Learning, PMLR, 2019, pp. 3763--3772.

\bibitem{arazo2019unsupervised}
E.~Arazo, D.~Ortego, P.~Albert, N.~O’Connor, K.~Mcguinness, Unsupervised
  label noise modeling and loss correction, in: International Conference on
  Machine Learning, 2019, pp. 312--321.

\bibitem{zhang2019addressing}
S.~Zhang, M.~Bansal, Addressing semantic drift in question generation for
  semi-supervised question answering, arXiv preprint arXiv:1909.06356, 2019.

\bibitem{yi2019probabilistic}
K.~Yi, J.~Wu, Probabilistic end-to-end noise correction for learning with noisy
  labels, in: Proceedings of the IEEE Conference on Computer Vision and Pattern
  Recognition, 2019, pp. 7017--7025.

\bibitem{li2019learning}
J.~Li, Y.~Wong, Q.~Zhao, M.~S. Kankanhalli, Learning to learn from noisy
  labeled data, in: Proceedings of the IEEE Conference on Computer Vision and
  Pattern Recognition, 2019, pp. 5051--5059.

\bibitem{bai2021understanding}
Y.~Bai, E.~Yang, B.~Han, Y.~Yang, J.~Li, Y.~Mao, G.~Niu, T.~Liu, Understanding
  and improving early stopping for learning with noisy labels, in: Advances in
  Neural Information Processing Systems, Vol.~34, 2021.

\bibitem{zhang2021FSR}
Z.~Zhang, T.~Pfister, Learning fast sample re-weighting without reward data,
  in: Proceedings of the IEEE/CVF International Conference on Computer Vision,
  2021, pp. 725--734.

\bibitem{nguyen2020tutorial}
L.~Nguyen, Tutorial on em algorithm, Tech. rep. (2020).

\bibitem{tanaka2018joint}
D.~Tanaka, D.~Ikami, T.~Yamasaki, K.~Aizawa, Joint optimization framework for
  learning with noisy labels, in: Proceedings of the IEEE Conference on
  Computer Vision and Pattern Recognition, 2018, pp. 5552--5560.

\bibitem{vgg}
K.~Simonyan, A.~Zisserman, Very deep convolutional networks for large-scale
  image recognition, International Conference on Learning Representations
  (ICLR), 2015.

\bibitem{huang2017densely}
G.~Huang, Z.~Liu, L.~Van Der~Maaten, K.~Q. Weinberger, Densely connected
  convolutional networks, in: IEEE conference on computer vision and pattern
  recognition, 2017, pp. 4700--4708.

\bibitem{ResNet18}
K.~He, X.~Zhang, S.~Ren, J.~Sun, Identity mappings in deep residual networks,
  in: European conference on computer vision, Springer, 2016, pp. 630--645.

\bibitem{han2019deep}
J.~Han, P.~Luo, X.~Wang, Deep self-learning from noisy labels, in: Proceedings
  of the IEEE International Conference on Computer Vision, 2019, pp.
  5138--5147.

\bibitem{deng2009imagenet}
J.~Deng, W.~Dong, R.~Socher, L.-J. Li, K.~Li, L.~Fei-Fei, Imagenet: A
  large-scale hierarchical image database, in: 2009 IEEE conference on computer
  vision and pattern recognition, Ieee, 2009, pp. 248--255.

\bibitem{chen2019understanding}
P.~Chen, B.~B. Liao, G.~Chen, S.~Zhang, Understanding and utilizing deep neural
  networks trained with noisy labels, in: International Conference on Machine
  Learning, PMLR, 2019, pp. 1062--1070.

\bibitem{lee2018cleannet}
K.-H. Lee, X.~He, L.~Zhang, L.~Yang, Cleannet: Transfer learning for scalable
  image classifier training with label noise, in: Proceedings of the IEEE
  Conference on Computer Vision and Pattern Recognition, 2018, pp. 5447--5456.

\end{thebibliography}

\newpage

\begin{minipage}{0.3\textwidth}
\begin{tikzpicture}[remember picture, overlay]
\node[text width=\linewidth]{
\begin{figure}[H]\centering
    \vspace{-2cm}
    \includegraphics[height=18cm, width=0.7\linewidth, keepaspectratio]{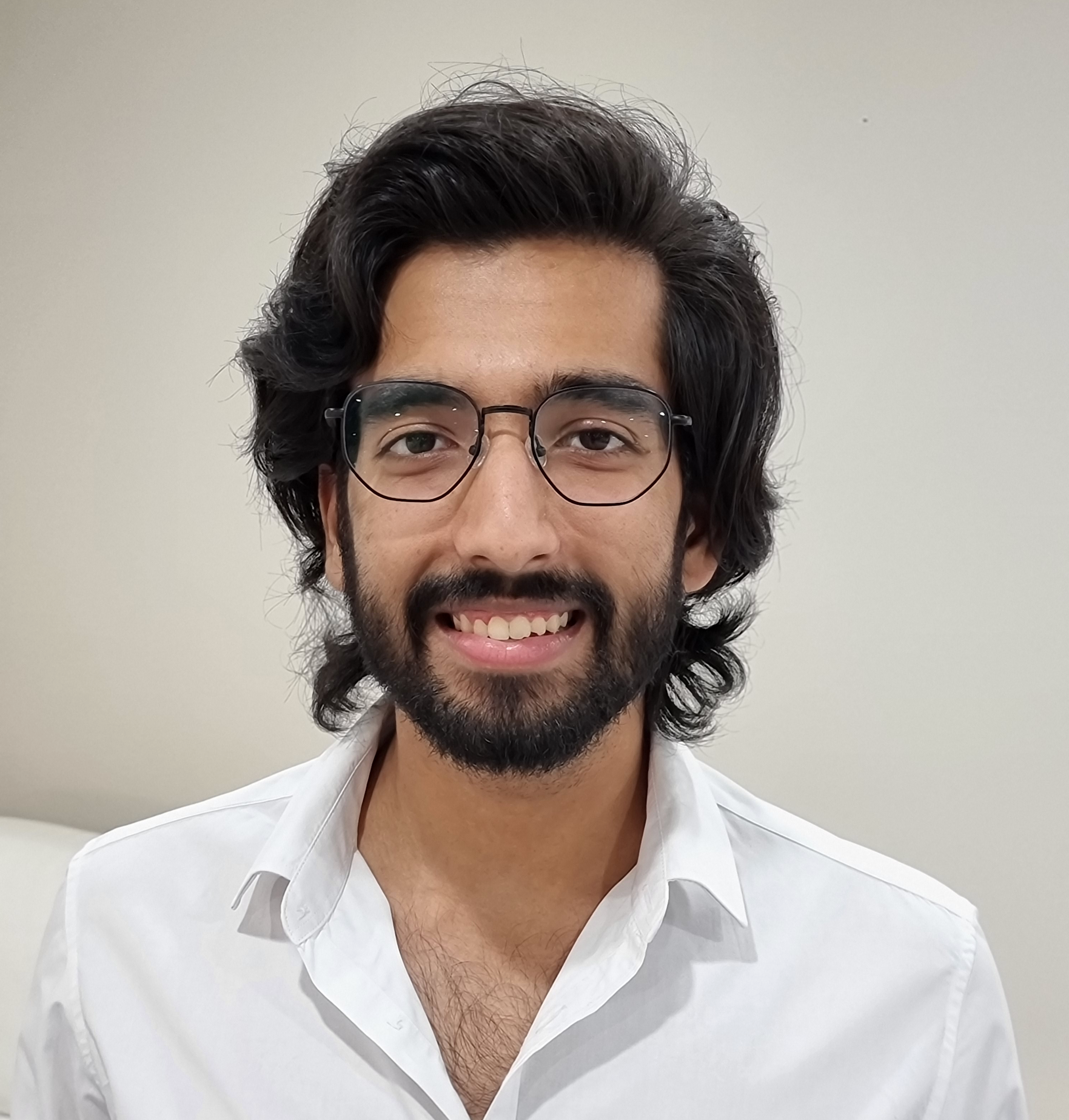}\\
    \vspace{1cm}
    \includegraphics[height=18cm, width=0.7\linewidth, keepaspectratio]{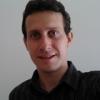}\\
    \vspace{1cm}
    \includegraphics[height=18cm, width=0.7\linewidth, keepaspectratio]{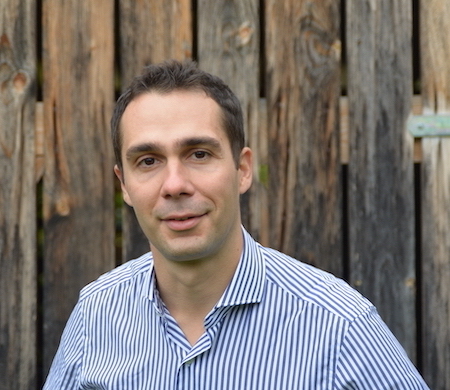}\\
    \vspace{2cm}
    \includegraphics[height=18cm, width=0.7\linewidth, keepaspectratio]{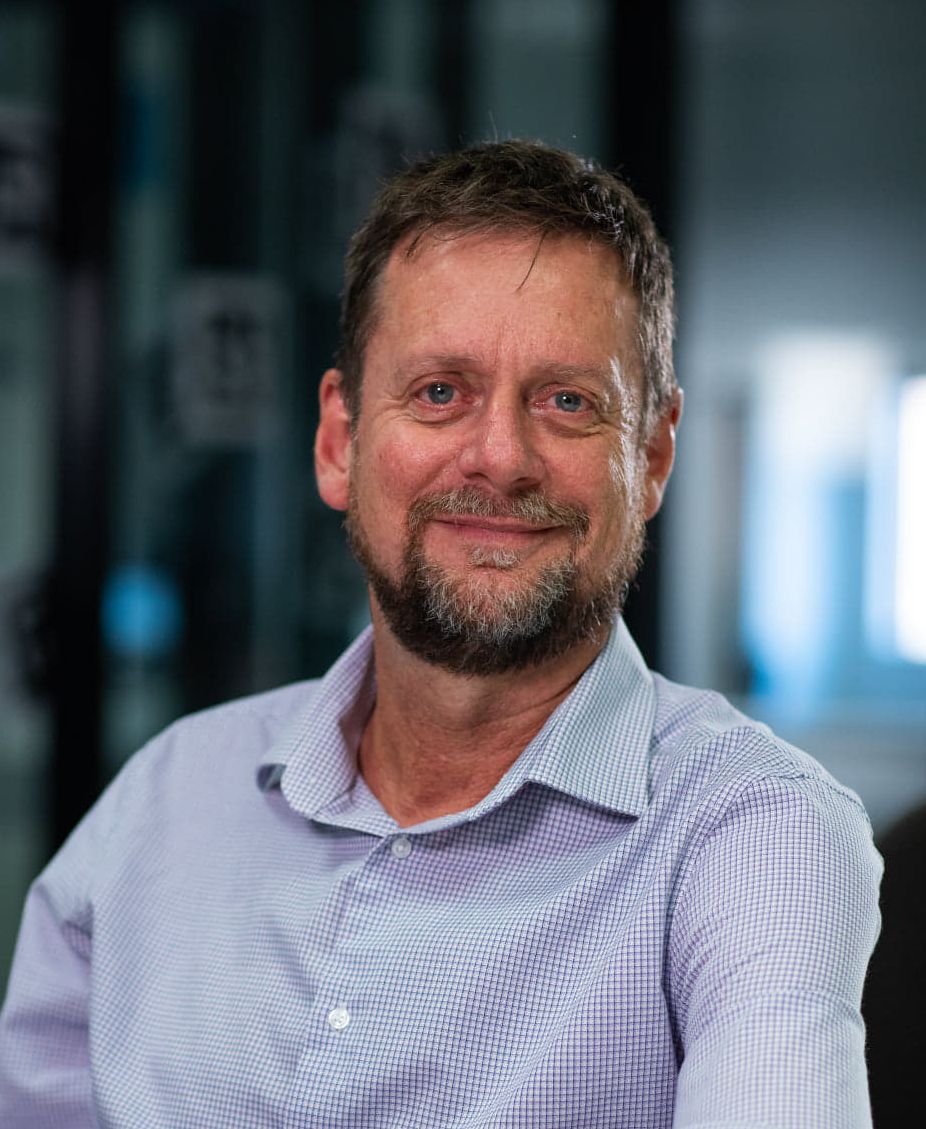}\\
    \vspace{1cm}
    \includegraphics[height=18cm, width=0.7\linewidth, keepaspectratio]{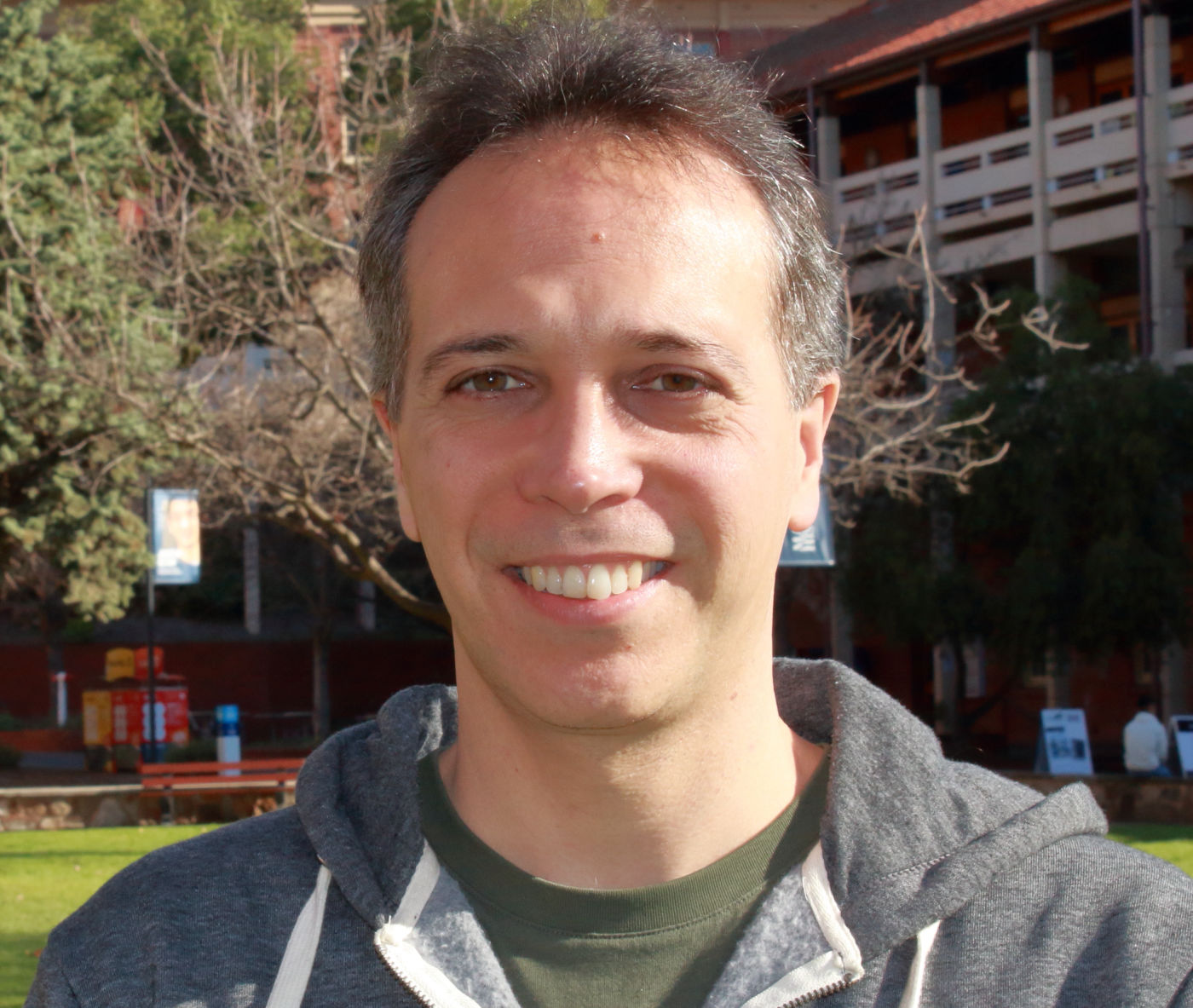}
\end{figure}
};
\end{tikzpicture}
\end{minipage}%
\hfill
\begin{minipage}{0.60\textwidth}\fontsize{8pt}{12pt}\selectfont
\begin{tabular}{p{\textwidth}}
\thispagestyle{empty}
Ragav Sachdeva is a PhD student in the Visual Geometry Group at the University of Oxford, supervised by Prof. Andrew Zisserman. He obtained his undergraduate degree in computer science at the University of Adelaide, where he did his honours thesis with Prof. Gustavo Carneiro.
\\[1\baselineskip]
Filipe R. Cordeiro is a professor of the Department of Computing at Universidade Federal Rural de Pernambuco (UFRPE). 
In 2015, he received his Ph.D. in computer science from the Federal University of Pernambuco (UFPE). 
Filipe's mains contributions are in the area of computer vision, medical image analysis, and machine learning.
\\[1\baselineskip]
Vasileios Belagiannis is a professor in the Faculty of Computer Science at Otto von Guericke University Magdeburg. His research deals with topics such as representation learning, uncertainty estimation, multi-modal learning, learning with different forms of supervision, learning algorithm for noisy labels, few-shot learning and meta-learning.
\\[1\baselineskip]
Ian Reid is the Head of the School of Computer Science at the University of Adelaide, and the senior researcher at the Australian Institute for Machine Learning. His research interests include robotic and active vision, visual tracking, SLAM, human motion capture and intelligent visual surveillance.
\\[1\baselineskip]
Gustavo Carneiro is a professor in the School of Computer Science at the University of Adelaide, Director of Medical Machine Learning at the Australian Institute of Machine Learning and an Australian Research Council Future Fellow. His main research interests are in computer vision, medical image analysis and machine learning. He is moving to the CVSSP at the University of Surrey in December 2022.

\end{tabular}
\end{minipage}%

\end{document}